\documentclass[10pt,twocolumn,letterpaper]{article}

\usepackage[pagenumbers]{cvpr} 

\usepackage[dvipsnames]{xcolor}
\definecolor{cvprblue}{rgb}{0.21,0.49,0.74}

\def\mymethod{TraF-Align}
 
\def\ie{\emph{i.e}\onedot} 
 
\def\etc{\emph{etc}\onedot}

\def\etal{\emph{et al}\onedot}

\usepackage{threeparttable}
\usepackage{bm}
\usepackage{colortbl}
\usepackage{booktabs}
\usepackage{helvet}
\usepackage[eulergreek]{sansmath}
\usepackage{multirow}
\usepackage{pgfplots}
\usetikzlibrary{spy}
\usepgfplotslibrary{groupplots}
\pgfplotsset{compat = 1.18,
    tick label style = {font=\sansmath\sffamily\scriptsize},
    every axis label = {font=\sansmath\sffamily\scriptsize},
    legend style = {font=\sansmath\sffamily\scriptsize},
    label style = {font=\sansmath\sffamily\scriptsize}
}
\usetikzlibrary{patterns}

\usepackage{graphicx}
\usepackage{flushend}

\definecolor{cvprblue}{rgb}{0.21,0.49,0.74}
\usepackage[pagebackref,breaklinks,colorlinks,allcolors=cvprblue]{hyperref}

\def\paperID{*****} 
\def\confName{CVPR}
\def\confYear{2025}

\title{\mymethod: Trajectory-aware Feature Alignment for Asynchronous \\ Multi-agent Perception}

\author{Zhiying Song\textsuperscript{1}, Lei Yang\textsuperscript{1}, Fuxi Wen\textsuperscript{1,2,$\dag$}, Jun Li\textsuperscript{1,2}\\
\textsuperscript{1}School of Vehicle and Mobility, Tsinghua University 
\textsuperscript{$\dag$}Corresponding author 
\\
\textsuperscript{2}State Key Lab of Intelligent Green Vehicle and Mobility \\
{\tt\small 
\{song-zy24,yanglei20\}@mails.tsinghua.edu.cn, \{wenfuxi,lijun1958\}@tsinghua.edu.cn}
}
\usepackage[accsupp]{axessibility}

\begin{document}
\maketitle

\begin{abstract}
Cooperative perception presents significant potential for enhancing the sensing capabilities of individual vehicles, however, inter-agent latency remains a critical challenge. Latencies cause misalignments in both spatial and semantic features, complicating the fusion of real-time observations from the ego vehicle with delayed data from others. To address these issues, we propose \mymethod, a novel framework that learns the flow path of features by predicting the feature-level trajectory of objects from past observations up to the ego vehicle’s current time. By generating temporally ordered sampling points along these paths, \mymethod \ directs attention from the current-time query to relevant historical features along each trajectory, supporting the reconstruction of current-time features and promoting semantic interaction across multiple frames. This approach corrects spatial misalignment and ensures semantic consistency across agents, effectively compensating for motion and achieving coherent feature fusion. Experiments on two real-world datasets, V2V4Real and DAIR-V2X-Seq, show that \mymethod \ sets a new benchmark for asynchronous cooperative perception. The code is available at \href{https://github.com/zhyingS/\mymethod}{https://github.com/zhyingS/\mymethod}.
\end{abstract}
    
\section{Introduction}
\label{sec:intro}
\begin{figure}
    \centering
    \includegraphics[width=0.97\linewidth]
    {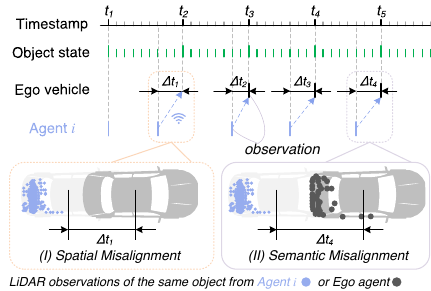}
    \caption{Asynchronous observations of ego vehicle and other agents result in spatial and semantic misalignments.}
    \label{fig:introduction}
\end{figure}

Intelligent vehicles rely on advanced onboard sensors, such as LiDAR and cameras, to perceive and understand the surrounding world. Nevertheless, the limited field of view and detection range of onboard sensors restrict their ability to perceive objects in occluded areas and at greater distances \cite{caillot2022survey,zhang2021emp,qiu2022autocast,xu2022opv2v}. Vehicle-to-everything (V2X) communication-enabled cooperative perception systems bridge this gap by exchanging  perception information between vehicles and other agents, thereby extending the perception capability of a single agent \cite{
chen2019cooper,chen2019f,yang2023bevheight,luextensible,zhang2024ermvp}.

In real-world scenarios, cooperative perception systems are cursed with inter-agent time delays \cite{wang2020v2vnet,xu2022v2x}. Consider a network with an ego agent and agent \textit{i}, where their observations of the same object are shown in Figure \ref{fig:introduction}. Latency in agent \textit{i}'s message, captured before ego's observation time, can result in either a complete loss of features (\ref{fig:introduction}-I) or feature misalignment relative to ego's observation (\ref{fig:introduction}-II). 
Corresponding to Figure \ref{fig:introduction}-I and \ref{fig:introduction}-II, there are two main challenges: i) Spatial misalignment, where observations from other agents fail to align with the ground truth in space, lacking any reference from the ego vehicle's observations. ii) Semantic misalignment, where both the ego vehicle and another agent observe the same object but fail to recognize it as the same entity.
These misalignments lead to mislocated objects and increased false positives \cite{song2023cooperative,song2024spatial}.

To address the first challenge, some studies have proposed potential solutions,
by predicting the current-frame features using historical frames \cite{leisyncnet}, or predicting the pixel-level flow of features and then extrapolating them \cite{yu2024flow} or directly warping them to their future positions \cite{NEURIPS2023_5a829e29}.  For instance, FFNet \cite{yu2024flow} uses a self-supervised approach to learn the first-order derivatives of features between consecutive frames, and CoBEVFlow \cite{NEURIPS2023_5a829e29} generates an instance-level region of interest (ROI) and calculates feature motion vectors by matching and tracking these regions. However, predicting the movement of high-dimensional neural features is non-trivial. Despite their efforts, these methods struggle with larger delays and lack end-to-end capability.
For the second challenge, the significance of semantic misalignment has not been well investigated in previous work.

To address these misalignments, our core idea is to \textit{reconstruct current-time features and achieve semantic consensus across agents on multi-frame LiDAR features} within a unified framework. We guide the flow and interaction of high-dimensional features by learning the motion of low-dimensional trajectory features. 
A trajectory field is defined as a spatiotemporal map of each object's anticipated path, guiding the generation of temporally ordered sampling points along each trajectory. Cross-attention between each query and its aligned sampling points allows the query to access relevant past features along the trajectory, enabling effective feature reconstruction and semantic interaction for robust spatial alignment and inter-agent consensus.
Consider the spatial misalignment scenario shown in Figure \ref{fig:introduction}-I. By aggregating multi-frame point cloud features from agent $i$, we learn the object's past trajectory distribution and predict its motion direction up to the ego vehicle's current time, forming a trajectory field. This field guides attention from the predicted position to historical features along the same trajectory, enabling the reconstruction of the object’s current features near the ground truth and compensating for motion.
For semantic misalignment cases (Figure \ref{fig:introduction}-II), while delays may misalign agent $i$'s observations with the ego vehicle's, the continuity of object motion implies overlap with point clouds from past ego frames. By learning the trajectory distribution from the ego’s multi-frame point clouds and applying attention to compute feature correlations along the trajectory,  the model recognizes that all features along the trajectory correspond to the same object, thus achieving cross-agent semantic alignment.

\begin{figure*}[t]
    \centering
    \includegraphics[width=0.98\linewidth]{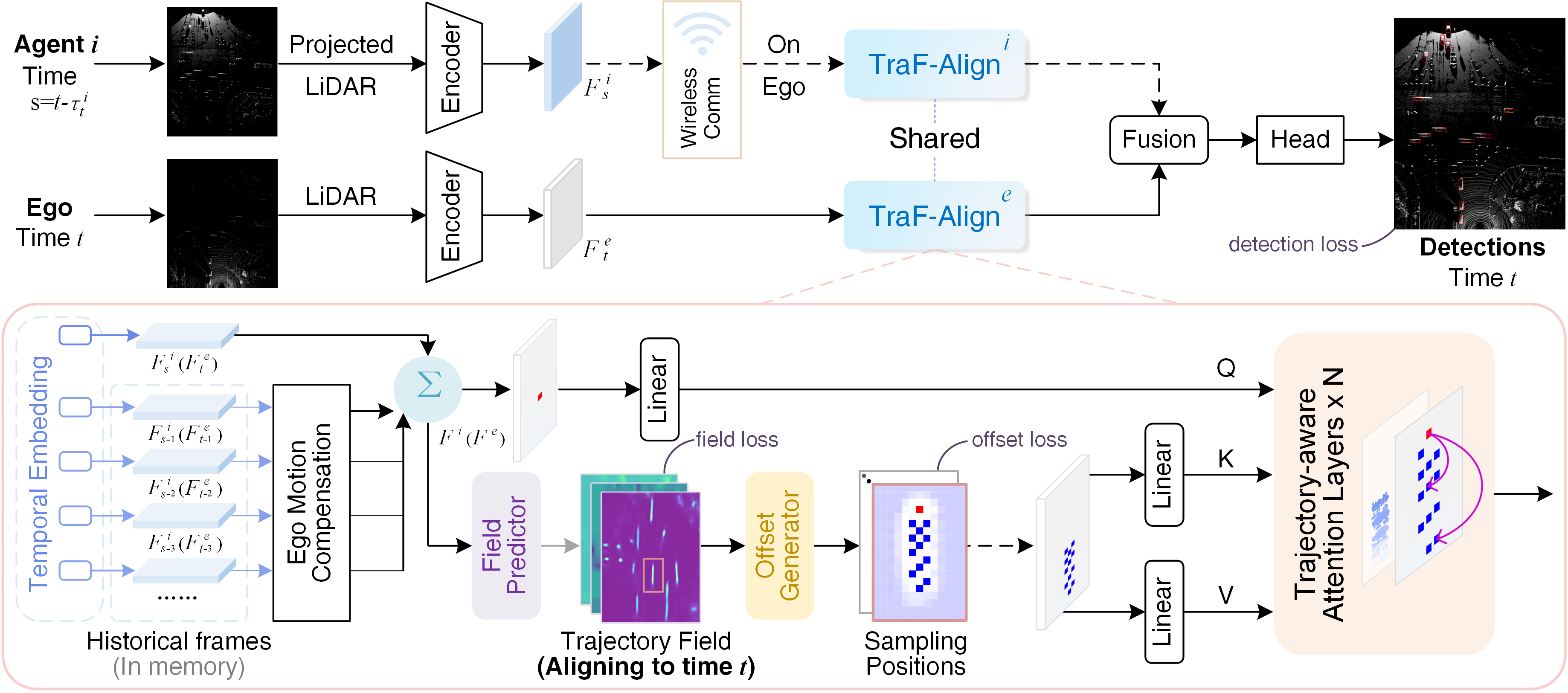}
    \caption{\textbf{Proposed Architecture.} With asynchronous LiDAR inputs from the ego agent (time $t$) and agent $i$ (time $s = t - \tau_t^i$), BEV features are extracted using an onboard sparse encoder. Agent $i$'s features are transmitted to the ego agent and stored in memory. \textbf{\mymethod} predicts a trajectory field up to time $t$, generates target attention positions, and uses features at these positions as keys/values in attention layers. Finally, the multi-agent features are fused and processed by the head to generate cooperative predictions at time $t$.}
    \label{fig:framework}
\end{figure*}

Building on the abovementioned idea, we propose \mymethod, a novel framework for asynchronous multi-agent cooperative perception (as shown in Figure \ref{fig:framework}). The proposed method includes three key parts: i) a field predictor that generates synchronized trajectory fields aligned with the ego vehicle's time, ii) an offset generator that produces sampling points for each query position and distributes them along the corresponding trajectory, and iii) attention layers where each query position is treated as a token, with attention focused only on positions specified by the offset generator.
Extensive experiments are conducted on two widely used large-scale real-world datasets, V2V4Real \cite{xu2023v2v4real} and DAIR-V2X-Seq \cite{yu2023v2x}, one for vehicle-to-vehicle (V2V) cooperative perception, another for vehicle-to-infrastructure (V2I) applications. Experimental results show that state-of-the-art (SOTA) performance is achieved for \mymethod \ in both object detection precision and robustness to inter-agent latency. 
The contributions are summarized as follows:
\begin{itemize}
    \item We propose \mymethod, a novel framework for asynchronous cooperative perception, addressing spatial and semantic misalignments caused by time delays.
    \item  We leverage a field predictor and an offset generator to align trajectory fields and distribute attention points along object trajectories, enabling robust feature reconstruction and interaction along spatiotemporally aligned paths.
    \item We introduce a field loss and an offset loss to supervise trajectory alignment and the generation of attention points, providing an end-to-end trainable solution.
\end{itemize}
\section{Related work}
\label{sec:related_work}

\noindent{\textbf{Asynchronous V2X cooperative perception.}} V2X cooperative perception extends the sensing range of single vehicles by exchanging information with other agents via wireless communication \cite{huang2023v2x}. Previous studies have conducted extensive research under time-synchronized conditions, such as F-Cooper \cite{chen2019f}, V2VNet \cite{wang2020v2vnet}, V2X-ViT \cite{xu2022v2x,xu2024v2x}, ERMVP \cite{zhang2024ermvp}, \etc. In the presence of latency, previous research emphasizes using historical data to estimate and compensate for object motion.
For instance, at the output level, the object
state filters are designed to predict and update the dynamic state
of the received objects \cite{cai2023consensus}.
Using raw data, Zhang \etal \cite{zhang2023robust} and Dao \etal \cite{dao2024practical} estimated the motion of point clouds and predicted compensatory flows for late-arriving data. 
However, these levels may encounter challenges, such as short prediction time horizons or high communication data volumes.
At the feature level, which has a better balance between accuracy
and transmission bandwidth \cite{liu2023towards}, SyncNet \cite{leisyncnet} employs a pyramidal LSTM to iteratively process and learn from delayed features, enabling the prediction of current-frame features.
FFNet \cite{yu2024flow} predicts inter-frame feature changes in a self-supervised feature manner to synchronize features across agents. Nevertheless, predicting high-dimensional features is challenging and prone to uncontrollable variations. 
Additionally, CoBEVFlow \cite{NEURIPS2023_5a829e29} extracts and warps object-specific features to anticipated locations by predicting instance-wise motion vectors. However, this approach requires generating ROIs and tracking their motion, resulting in a non-end-to-end two-stage method. This paper uses multi-frame LiDAR sequences to predict low-dimensional trajectory features and guide attention to fuse features on past trajectories.
By operating directly at the feature level, our approach enables end-to-end training and avoids the complexities of high-dimensional feature prediction.

\noindent{\textbf{Learning from LiDAR sequences across agents.}}
Multi-frame point cloud fusion is widely applied in single-vehicle autonomous systems to amalgamate semantic data across different perspectives \cite{wang2024sequential,yang20213d,qi2021offboard,ma2023detzero,chen2022mppnet,rong2023dynstatf}. However, its application in V2X cooperative perception remains under-explored. Although studies like SCOPE \cite{yang2023spatio} and MRCNet \cite{hong2024multi} have leveraged recurrent neural networks to fuse sequences from ego vehicles, these methods are resource-intensive and limited by sequence length. This has restricted their use to ego vehicles rather than a broader agent network, necessitating a more straightforward multi-frame fusion network.
In cooperative perception scenarios, multi-frame point clouds offer crucial motion cues that enable agents to synchronize their understanding of object positions despite time delays, thereby correlating trajectories across agents.  
This aspect has yet to be fully explored at the feature level. This paper proposes a novel approach using trajectory fields and customized attention with offsets for direct multi-frame fusion, thus establishing a consistent understanding of object motion across the network.
\section{Method}

 \subsection{Problem formulation}
Consider a group of networked agents indexed by 
$\mathcal{I}=\{e,1,2,\dots,N-1\}$, each is equipped with LiDAR and communication units. Index $e$ indicates the ego agent, which integrates information from other agents to extend its sensing range and reduce the blind spots. The LiDAR observation of the $i$th agent at its local timestamp $s$ is denoted as $L_s^i$, where $i\in \mathcal{I}$. A shared neural network $f(\cdot)$ processes this observation into deep features. At each timestamp $s$, agent $i$ transmits $F_s^i=f(L_s^i)$ to the ego agent, who then accumulates these features in the cache, 
forming a historical feature set for agent $i$ at time $s$, represented as $\mathcal{F}_s^i = \{F_{s-k}^i\}_{k\in K}$, where $K=\{k\in \mathbb{Z}|1 \leqslant k \leqslant m_i\}$, $m_i$ is the maximal sequence length for agent $i$. 
Accounting for delays in the collaborative process, messages from various agents arrive at ego at different times. Let $\tau^i_t$ be the delay for agent $i$'s message, and this means the information sent by agent $i$ at time $t-\tau_t^i$ is received by the ego at time $t$. If ignoring feature extraction delay, then $s=t-\tau_t^i$.
The task is to fuse these multi-agent features and estimate the state of objects, \ie, the bounding boxes, at ego timestamp $t$ within the detection range. The information available for fusion includes: i) the ego's historical features $\mathcal{F}_t^e$, and ii) the historical features from other agents, denoted as $\{\mathcal{F}^i_{t-\tau_t^i}\}_{ i\in \mathcal{I} \backslash \{e\} }$. The proposed method is shown in Figure \ref{fig:framework}.

\subsection{Feature extraction}
The encoder converts single-frame raw point cloud data into BEV features. Similar to  PointPillars \cite{lang2019pointpillars}, the point cloud is first discretized into a regular grid on the $x$-$y$ plane.
Within each pillar, the point features are encapsulated as the vector $[x,y,z,x_c,y_c,z_c,x_p,y_p,z_p] \in \mathbb{R}^{9}$. Here, $[x,y,z]$ is the point's position, while $[x_c,y_c,z_c]$ indicates the center of the pillar. The vector $[x_p,y_p,z_p]$ represents the point's offset from the pillar center.
The pillar feature extraction and backbone network architectures are adopted from PillarNext \cite{li2023pillarnext}. Multi-Layer Perceptron (MLP) serves to encode point-wise features, while a sparse variant of ResNet \cite{he2016deep} is utilized as the backbone to distill deeper and more complex features with size $(C,H,W)$. The sparse network is implemented using Spatially Sparse Convolution (SpConv) library \cite{spconv2022}.
The extracted single-frame features are transmitted and stored in the ego's feature bank ($\mathcal{F}_t^e$ and $\mathcal{F}_s^i$), which maintains a fixed capacity by admitting new features and discarding the oldest ones, ensuring an up-to-date repository for feature reuse. The extracted features of each agent are then processed using \mymethod \ separately. 

\noindent \textbf{Temporal Encoding.} 
Temporal information is decoupled from the point features to facilitate the reuse of historical features across frames. This allows for the immediate retrieval of historical features at any timestamp.
Inspired by the positional embedding in Transformer \cite{vaswani2017attention}, a temporal embedding $TE\in \mathbb{R}^{C\times H \times W}$ is defined as 
\begin{equation}
\label{eq:temporal_encoding}
TE_{(\tau, 2j)}  =\sin (\xi), \  TE_{(\tau, 2j+1)}  =\cos (\xi), 
\end{equation}
where $\xi=\tau / \varepsilon^{2j / C}$,
$j$ indexes the dimension along $C$ channels, $\tau$ represents the number of frames delayed from the current frame, and  $\varepsilon$ is set to eight in this study.
Each feature within the memory cache,  $\mathcal{F}_t^e$ for the ego agent and $\mathcal{F}_s^i$ for agent $i$, associated with a specific temporal delay, is concatenated with a corresponding temporal embedding. These concatenated features are refined through a convolutional layer with a $1 \times 1$ kernel and then warped to time $t$ by ego-motion compensation.

\subsection{Trajectory field} 
Each agent processes its aggregated multi-frame feature map via a field predictor, which generates a trajectory field consisting of position and orientation fields. The position field creates heatmap peaks along the object’s trajectory, while the orientation field captures the inverse tangent direction of the trajectory.
Due to delays, agents' local observations may be outdated. To achieve temporal alignment, the field predictor predicts the trajectory features of all objects up to the ego vehicle’s latest time (with temporal information encoded in the features via Equation \ref{eq:temporal_encoding}). This ensures that, for the same object, trajectory fields from different viewpoints spatially overlap and are aligned temporally to the ego vehicle's current time.
Our model uses a variant of UNet \cite{ronneberger2015u} as the field predictor, whose architecture is detailed in the supplementary material, along with a supervision signal to guide the learning process.

\begin{figure}[t]
    \centering
    \includegraphics[width=0.99\linewidth]{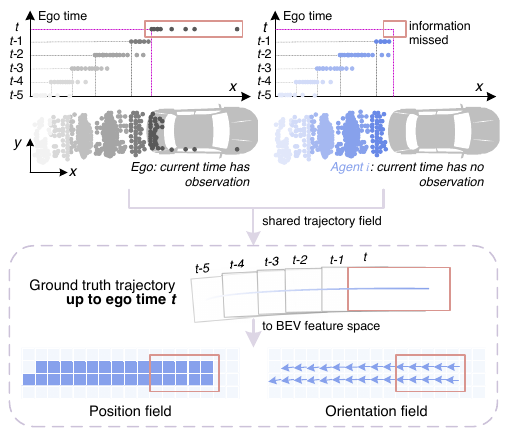}
    \captionsetup{justification=raggedright,singlelinecheck=false}
    \caption{Illustration of trajectory field, which includes a position field indicating the occupancy grid of the trajectory and a direction field depicting the trajectory's flow orientation.}
    \label{fig:traj_field}
\end{figure}

As shown in Figure \ref{fig:traj_field}, when generating the ground truth field,  we first extract bounding box annotations of each object from the local perspective of each agent within the maximum temporal window, starting from the object's initial appearance up to the ego's current time.  A trajectory is then constructed by connecting the centers of these bounding boxes. Specifically, the length of the object's trajectory is given by 
$    \lceil \tau_i \cdot \omega_i \rceil + m_i, \ \forall \ i \in \mathcal{I}
$,
where $\lceil \cdot \rceil$ denotes the ceiling function, $\omega_i$ is the LiDAR sampling frequency (s$^{-1}$) of agent $i$, and $\tau_i$ represents its cooperative latency (s).
The trajectory field is then generated by interpolating along the trajectory and projecting it onto the BEV feature space. If two object trajectories overlap, the trajectory with the more recent timestamp will overwrite the older one. 

For cases in Figure \ref{fig:introduction}-I, such a design compels the network to predict the trajectory field up to the present moment, thereby supporting the learning of motion information.  Failing to do so would result in the network detecting only the delayed bounding box. Compared to the direct forecasting of high-dimensional features from the backbone network, predicting the trajectory field provides a more straightforward and more reliable method with predictable outcomes.

\subsection{Trajectory-aware attention} 
\begin{figure}[t]
    \centering
    \subfloat[Trajectory-aware attention layers.]{
    \label{fig:offset_a}
    \includegraphics[width=0.97\linewidth]{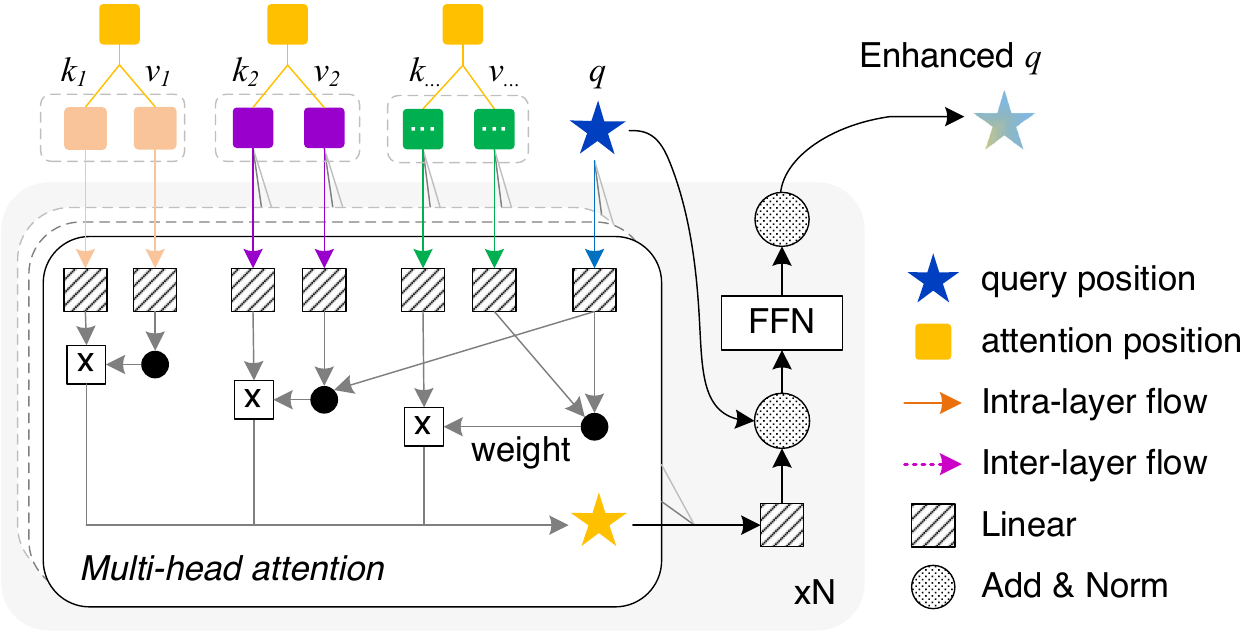}}
    \\
    \vspace{0.1cm}
    \subfloat[Query and ground truth attention positions.]{
    \label{fig:offset_b}
    \includegraphics[width=0.94\linewidth]{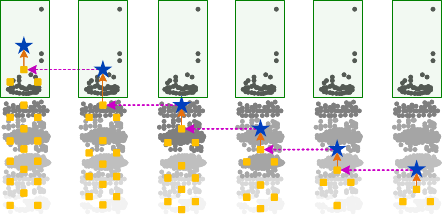}}
    \caption{Attention positions and trajectory-aware attention layers.}
    \label{fig:offset}
    
\end{figure}

We flatten the input feature map $F^i$ ($\forall i \in \mathcal{I}$) of size $C \times H \times W$ into $H \times W$ embeddings, each serving as a query $q \in \mathbb{R}^{1 \times C}$. For each query, the offset generator creates a response set $R \in \mathbb{R}^{n \times C}$, capturing features at designated attention positions that are learned based on predefined ground truth. $n$ is the number of attention positions for each query, and it's set to $18$ in this paper. 
To generate the ground truth offsets, we select $n$ pixels on the feature map that satisfy three conditions: i) The position field has a response (indicating the presence of an object), ii) The pixels belong to the same trajectory as the query (indicating the same object), and iii) The pixels correspond to earlier timestamps than the query (indicating historical positions of the object).
As shown in Figure \ref{fig:offset_b}, this ground truth aligns each query with preceding trajectory positions, enabling it to draw in relevant historical information. The trajectory field has effectively modeled the motion and flow of the historical features, offering valuable guidance to the offset generator. With this rich information, the generator can rely on simple convolution layers to produce good attention offsets, and we use PReLU activation \cite{He_2015_ICCV} to provide flexibility in adjusting the offsets’ directional orientation.
Given the response set $R$, the feature of query $q$ is updated or reconstructed by the attention process:
\begin{equation}
    \text{Attention}\left( q, R\right) = \text{Softmax}\left[ \frac{qW^q  \left(  RW^k\right)^T}{\sqrt{d }}\right] RW^v, 
\end{equation}
where $W^q, W^k, W^v \in \mathbb{R}^{C\times d}$ are the query, key and value projection matrices. Similar to the original Transformer \cite{vaswani2017attention}, we also incorporate feed-forward networks, addition and normalization layers, and a multi-head attention design, the details are shown in Figure \ref{fig:offset_a}. Using attention along the trajectory transfers features to the current time step and enables feature interaction along the trajectory. This allows the model to recognize which features belong to the same object, aligning the features across agents for effective fusion. As a result, each agent aligns its feature map to the ego vehicle's latest time step.

\subsection{Inter-agent fusion, detection head and loss}
\label{sec:head_loss}
While inter-agent fusion has been extensively studied \cite{chen2019f,xu2022opv2v,zhang2024ermvp}, this paper adopts a simplified approach by concatenating features from all agents and utilizing a basic convolution layer for fusion.
We use an anchor-based detection head and employ three losses: a detection loss, a field loss, and an offset loss:
\begin{equation}
    \mathcal{L} = \mathcal{L}_{\text{detection}} + \alpha \mathcal{L}_{\text{field}} + \beta \mathcal{L}_{\text{offset}},
\end{equation}
where $\mathcal{L}_{\text{detection}}$ is consistent with the loss used in \cite{lang2019pointpillars}, $\mathcal{L}_{\text{field}}$ includes a focal loss \cite{lin2017focal} for the position field and an $L1$ loss for the orientation field. The constants $\alpha$ and $\beta$ balance the contributions of each loss term. Here, we set $\alpha = \beta = 0.05$.
Considering the offset loss, because the generated attention positions are unordered, we first apply the Sinkhorn algorithm \cite{sinkhorn1967concerning,cuturi2013sinkhorn,sarlin2020superglue} to compute the matching probabilities $P_{jk}\in \mathbb{R}^{n \times n}$ between the generated attention positions and the ground truth. This is done using the L1 distance matrix $C_{jk}\in \mathbb{R}^{n \times n}$ as the cost for the pixel located at $(j,k)$ on the feature map. The offset loss is then defined as:
\begin{equation}
    \mathcal{L}_{\text{offset}} = \sum_{j,k} P_{jk} \cdot C_{jk}, \ \forall j \in [1,H], k\in[1,W].
\end{equation}

\section{Experiments}
To evaluate robustness to cross-agent latency, we conduct 3D object detection experiments on two large-scale real-world datasets, V2V4Real \cite{xu2023v2v4real} and DAIR-V2X-Seq \cite{yu2023v2x}. These datasets encompass both V2V and V2I scenarios, as well as sparse and dense LiDAR conditions. The detection results are evaluated using Average Precision (AP) at Intersection-over-Union (IoU) thresholds of $0.5$ (AP50) and $0.7$ (AP70).
To enable a fair comparison, we omit feature compression operations from the methods under evaluation.

\begin{table*}[t]
\footnotesize
\centering
\caption{Performance comparison of \textit{vehicle} class on V2V4Real validation and DAIR-V2X-Seq validation dataset with different latencies. The results are reported in AP $ 50/70$ (\%).}
\definecolor{color}{RGB}{255, 0, 0}   
\setlength{\tabcolsep}{2.2pt} 
\renewcommand\arraystretch{1.1} 

\begin{tabular}{c|ccccc|ccccc}
    \toprule
    \multirow{2}{*}{{Method}} &\multicolumn{5}{c|}{{V2V4Real}} & \multicolumn{5}{c}{{DAIR-V2X-Seq}}  \\
    \cline{2-11}
    &{0ms}&{100 ms}&{200 ms}&{300 ms}&{400 ms}&{0ms}&{100 ms}&{200 ms}&{300 ms}&{400 ms}\\
    \hline  
    Single Agent &48.75/30.53&48.75/30.53&48.75/30.53&48.75/30.53&48.75/30.53&51.31/40.45&51.31/40.45&51.31/40.45&51.31/40.45&51.31/40.45 \\
    AttFuse \cite{xu2022opv2v}&63.84/33.30&59.16/27.36&55.57/26.03&53.25/25.44&51.69/25.09&68.94/49.68&65.83/45.76&63.02/44.24&61.61/43.28&60.19/42.53\\
    F-Cooper \cite{chen2019f}&66.86/33.39&62.15/27.61&58.03/26.30&55.31/25.46&53.92/24.95&69.47/50.01&65.85/44.96&62.74/43.21&61.12/42.28&60.01/41.70\\
     V2VNet \cite{wang2020v2vnet} &67.11/35.47&61.97/29.14&58.09/27.85&56.26/27.43&54.62/26.75&71.44/51.91&67.26/46.63&63.84/44.59&61.68/43.24&59.78/42.08
\\    
     %
      {Where2comm \cite{hu2022where2comm}}&67.03/37.55&61.86/31.52&58.22/30.20&56.22/29.71&54.47/29.07&68.37/48.50&64.81/44.32&61.51/42.80&59.66/41.63&58.25/40.80 \\
      V2X-ViT \cite{xu2022v2x} &68.51/35.88&63.45/29.57&59.42/28.04&57.35/27.65&55.60/27.20&70.21/50.92&66.42/46.44&62.58/44.29&61.00/43.56&59.63/42.66\\    
    MRCNet \cite{hong2024multi} &67.83/37.37&62.65/31.59&58.98/30.39&56.70/29.53&55.03/28.94&69.27/50.67&65.90/45.60&62.65/44.15&60.09/42.39&58.19/41.35\\
      CoBEVT \cite{xu2023cobevt} &70.59/38.82&65.50/32.11&61.55/30.40&58.91/29.54&57.05/29.26&71.51/51.31&67.60/46.55&64.38/45.03&62.39/44.11&60.74/43.08\\
      ERMVP \cite{zhang2024ermvp} &70.74/38.45&65.53/31.91&61.63/30.53&59.24/30.18&57.26/29.31&71.19/{53.85}&67.04/48.09&63.67/46.37&61.23/44.76&59.55/43.93\\
    {\mymethod \ (Ours)} &\textbf{74.28/44.14}&\textbf{73.04/39.99}&\textbf{71.51/39.54}&\textbf{70.07/38.99}&\textbf{69.41/38.49}&\textbf{76.90/58.31}&\textbf{73.95/53.68}&\textbf{72.69/52.83}&\textbf{71.66/52.33}&\textbf{71.22/52.01}\\
\bottomrule
\end{tabular}

\label{tab:v2x-seq_0ms}
\end{table*}

\subsection{Implementation details}
\noindent{\textbf{Datasets and Benchmarks.}} 
V2V4Real \cite{xu2023v2v4real} is specialized for V2V cooperative detection in urban settings, and DAIR-V2X-Seq \cite{yu2023v2x} captures sequential data in V2I scenarios at intersections. The benchmarks include single-agent perception relying on the ego vehicle's data for detection, alongside eight SOTA intermediate fusion models: F-Cooper \cite{chen2019f}, V2VNet \cite{wang2020v2vnet}, Where2comm \cite{hu2022where2comm}, AttFuse \cite{xu2022opv2v}, V2X-ViT \cite{xu2022v2x}, CoBEVT \cite{xu2023cobevt}, MRCNet \cite{hong2024multi}, and ERMVP \cite{zhang2024ermvp}. 

\noindent{\textbf{Training details.}}
All the models are trained within the same codebase for $60$ epochs on nine RTX 3090 GPUs, employing the one-cycle learning rate strategy \cite{smith2019super} with AdamW \cite{loshchilov2019decoupledweightdecayregularization} as the optimizer. The point clouds are segmented at a grid size of $0.4 \times 0.4$ m, and the backbone's feature map size output is one-fourth the original size. During training on V2V4Real, all models shuffle the vehicle list and randomly select the ego vehicle at the start of each epoch. Standard LiDAR data augmentations are applied, including random flip, scaling, rotation, and translation. Multi-frame methods are trained with random latency augmentation, following a uniform distribution from $0$ to $400$ ms. \mymethod \  leverages two LiDAR frames for the ego vehicle and four for the cooperative vehicle on V2V4Real, while using four LiDAR frames for the vehicle and infrastructure on DAIR-V2X-Seq. The number of attention layers is set to two, each containing four attention heads.

\noindent{\textbf{Latency settings.}} To comprehensively evaluate robustness to time delays, we introduce delays of $0$, $100$, $200$, $300$, and $400$ ms to one of the agents separately, extensively testing the model's performance under each delay condition. Unless specified otherwise, these delays are applied to the point cloud data of the non-ego agent.

\subsection{Comparison with state-of-the-art}
\noindent{\textbf{Detection performance.}} Table \ref{tab:v2x-seq_0ms} presents a comparison of detection performance on two real-world datasets for \textit{vehicle} class, which includes cars, vans, trucks, buses, \etc. It's observed that the proposed method i) achieves state-of-the-art performance across all benchmarks under various latency conditions on both datasets. ii) In synchronous scenarios, \ie, when the delay is $0$ ms, our model improves AP $50/70$ by $25.53\%/13.61\%$ (on V2V4Real) and $25.59\%/18.26\%$ (on DAIR-V2X-Seq) compared to single-agent perception, it also surpasses ERMVP, the SOTA performance of feature-level methods, by $3.54\%/5.69\%$ and $5.71\%/4.46\%$, respectively. 

\noindent{\textbf{Robustness to inter-agent latency.}}
As shown in Table \ref{tab:v2x-seq_0ms}, our model demonstrates stronger robustness in the presence of time delays. Even under a $400$ ms delay, our AP decreases by only $4.87\%/5.65\%$ (on V2V4Real) and $5.68\%/6.30\%$ (on DAIR-V2X-Seq) compared to the synchronous scenario. This performance significantly outperforms the ERMVP, which shows declines of $13.48\%/9.14\%$ and $11.64\%/9.92\%$, respectively. 

\begin{figure}[t]
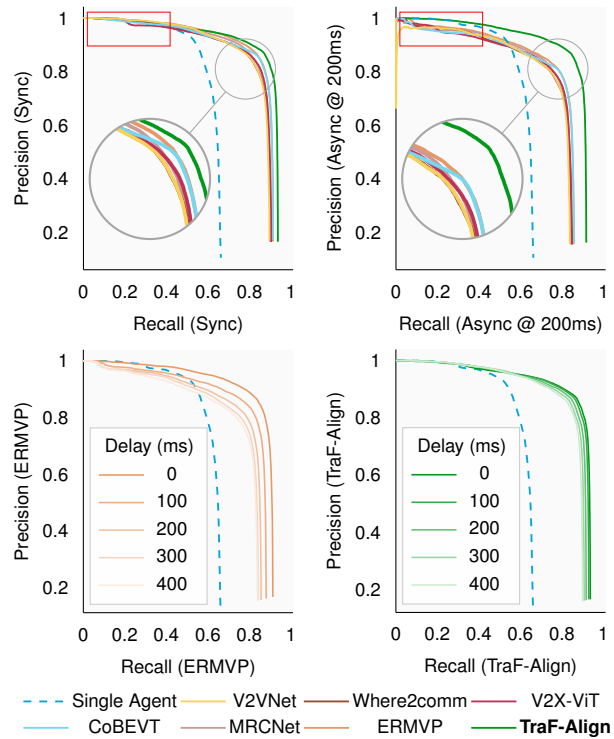

\centering
\subfloat{
\input{table/pr_curve/v2v4real_0ms_iou0.5}}
\subfloat{
\input{table/pr_curve/v2v4real_200ms_iou0.5}}
\\
\subfloat{
\input{table/pr_curve/v2v4real_mrcnet_iou0.5}}
\subfloat{
\input{table/pr_curve/v2v4real_defu_iou0.5}}
\vfill
\subfloat{
\definecolor{mycolor1}{rgb}{0.00000,0.65000,0.85000}
\definecolor{mycolor2}{rgb}{0.95000,0.60000,0.25000}
\definecolor{mycolor3}{rgb}{0.97000,0.82000,0.35000}
\definecolor{mycolor4}{rgb}{0.60000,0.30000,0.75000}
\definecolor{mycolor5}{rgb}{0,0.6,0.1}
\definecolor{mycolor6}{rgb}{0.45000,0.85000,0.95000}
\definecolor{mycolor7}{rgb}{0.75000,0.20000,0.35000}
\definecolor{mycolor8}{rgb}{0.8,0.6,0.6}
\definecolor{mycolor9}{rgb}{0.4,0.7,0.9}
\definecolor{mycolor10}{rgb}{0.7,0.6,0.4}
\definecolor{mycolor11}{rgb}{0.9,0.8,0.3}
\definecolor{mycolor12}{rgb}{0.5,0.4,0.7}
\definecolor{mycolor13}{rgb}{0.6,0.3,0.2}
\definecolor{mycolor14}{rgb}{0.9,0.6,0.4}
\definecolor{mycolor15}{rgb}{0.3,0.7,0.8}

\begin{tikzpicture}[font=\sffamily\scriptsize]
\node[draw=white,outer sep=0.5pt,inner sep=0pt] at (0,0) 
{
    \renewcommand{\arraystretch}{1.25} 
    \setlength{\tabcolsep}{0.5pt} 
     \begin{tabular}{cccccccc}
     \ref{prcurve_nofusion}&Single Agent &
     \ref{prcurve_v2vnet} & V2VNet&
      \ref{prcurve_where2comm} & Where2comm &
      \ref{prcurve_v2xvit} & V2X-ViT \\
     \ref{prcurve_cobevt} & CoBEVT & \ref{prcurve_mrcnet} & MRCNet &\ref{prcurve_ermvp}& ERMVP & \ref{prcurve_defu}& \textbf{\mymethod}\\
    \end{tabular}
    };
\end{tikzpicture}}
\caption{Precision-Recall curves for models on the V2V4Real validation dataset under various latency conditions.}
\label{experiments:robustness assessment}
\end{figure}

To gain a deeper understanding of the impact of latency on collaboration, we plot the Precision-Recall curves of the models under various latency conditions, as shown in Figure \ref{experiments:robustness assessment}. It is obvious that cooperation is effective in enhancing the recall limit of single-vehicle perception, helping the ego vehicle to detect a more significant number of objects. However, at a $200$ ms latency, state-of-the-art methods exhibit lower precision in the low-recall range compared to single-vehicle perception. This suggests that the delayed collaborative information interferes with the objects that the single vehicle could detect, corresponding to the \textit{semantic misalignment} issues depicted in Figure \ref{fig:introduction}-II. In areas not visible to the ego vehicle, specifically at high recall levels, the precision of SOTA methods significantly declines with the introduction of latency. This indicates that \textit{spatial misalignment} issues (Figure \ref{fig:introduction}-I) become dominant, leading to decreased performance. These findings reinforce our analysis of latency's impact on cooperative perception.
Furthermore, using ERMVP as an example, the differences between the PR curves diminish as latency increases, indicating that existing models progressively lose their effectiveness in managing delays. In contrast, \mymethod \ retains the original performance of single-agent perception in low-recall regions and shows no significant decline in performance in high-recall areas due to latency. This highlights our model's ability to establish strong semantic consistency between the ego vehicle's data and the received information, demonstrating robust reconstruction  capability and effectively addressing the scenarios in Figures \ref{fig:introduction}-I and \ref{fig:introduction}-II.

\begin{table}[b]
\footnotesize
\centering
\caption{Results of the ablation study of the core components on the DAIR-V2X-Seq dataset.}
\footnotesize
\definecolor{color}{RGB}{255, 0, 0}   
\renewcommand\arraystretch{1.1} 
\setlength{\tabcolsep}{0.75pt} 
\renewcommand\arraystretch{1.1} 
\begin{threeparttable}
\begin{tabular}{ccc|c|c|c|c|c}
    \toprule
    \multirow{2}{*}{FP}&\multirow{2}{*}{OG}&\multirow{2}{*}{AL}& \multicolumn{5}{c}{Latency of Infrastructure}  \\
    \cline{4-8}
    &&&0ms&100ms&200ms&300ms&400ms\\
    \hline  
     &&&75.35/54.72&73.01/50.81&71.88/50.17&71.04/49.77&70.09/49.45\\ %
     &&$\checkmark$&76.31/57.02&74.60/53.53&73.90/52.34&72.91/51.99&71.84/51.02\\
     &$\checkmark$&$\checkmark$&77.11/58.63&75.15/54.77&74.18/53.77&73.24/52.99&71.96/52.19
 \\

$\checkmark$&$\checkmark$&$\checkmark$&77.83/59.72	&75.74/55.85&74.99/54.82&74.10/54.07&72.89/53.21\\
\midrule
%
%
%
%
    \multirow{2}{*}{FP}&\multirow{2}{*}{OG}&\multirow{2}{*}{AL}& \multicolumn{5}{c}{Latency of Vehicle}  \\
        \cline{4-8}
    &&&0ms&100ms&200ms&300ms&400ms\\
    \hline  
     &&&75.35/54.72&74.31/49.61&72.17/43.32&67.19/34.82&56.93/24.51\\%
    &&$\checkmark$&76.31/57.02&75.10/51.93&72.91/46.22&69.22/39.15&59.93/27.87
\\
     &$\checkmark$&$\checkmark$&77.11/58.63&75.76/53.48&73.44/47.55&69.92/40.52&62.05/29.68
 \\    $\checkmark$&$\checkmark$&$\checkmark$&77.83/59.72&76.41/54.88&74.24/49.46&70.75/42.62&63.59/32.31 \\
\bottomrule
\end{tabular}
\begin{tablenotes}
\item FP: \textit{Field predictor}. OG: \textit{Offset generator}. AL: \textit{Attention layers}.
\end{tablenotes}
\end{threeparttable}

\label{tab:ablation_component}
\end{table}

\subsection{Ablation studies}
\label{sec:ablation}
In this section, we conducted ablation studies on the effectiveness of significant components and loss and multi-frame point clouds using average precision as the evaluation metric. Other ablation experiments and analyses (using PR curves) are presented in the supplementary material.

\noindent{\textbf{Effectiveness of Major Components.}} To evaluate the contribution of modules, we progressively remove i) the field predictor, ii) the offset generator, and iii) the attention layers. The results are shown in Table \ref{tab:ablation_component}. The experiments are conducted on the DAIR-V2X-Seq dataset. To ensure a comprehensive assessment of each component's contributions, we train the model for $120$ epochs, double the duration of the experiments presented in Table \ref{tab:v2x-seq_0ms}. Additionally, we test the effects of adding latency to both the ego and non-ego agents to fully evaluate the impact of each module. The consistent
rise in detection precision over different latency settings demonstrates the effectiveness of each introduced component. Notably, applying latency to vehicles results in lower AP than when latency is applied to infrastructure. This discrepancy arises from the different distributions of object visibility between the vehicle agent and roadside infrastructure agent in the dataset. 

\noindent \textbf{Effectiveness of loss.} Table \ref{tab:ablation_component_loss} presents the ablation study of field loss and offset loss, both designed to guide the attention biases, under the same settings as in Table  \ref{tab:ablation_component}. The results show that introducing the field loss provides better initial values for the offset generator, enhancing the model's performance. Optimal results are achieved only when both losses are incorporated.

\begin{table}[t]
\footnotesize
\centering
\caption{Results of the ablation study of the loss on the DAIR-V2X-Seq dataset.}
\footnotesize
\definecolor{color}{RGB}{255, 0, 0}   
\renewcommand\arraystretch{1.1} 
\setlength{\tabcolsep}{1.5pt} 
\renewcommand\arraystretch{1.1} 
\begin{threeparttable}
\begin{tabular}{cc|c|c|c|c|c}
    \toprule
    \multirow{2}{*}{FL}&\multirow{2}{*}{OL}& \multicolumn{5}{c}{Latency of Infrastructure}  \\
    \cline{3-7}
    &&0ms&100ms&200ms&300ms&400ms\\
    \hline  
    &&76.43/57.74&74.44/53.91&73.65/52.97&72.71/52.10&71.83/51.87\\
          $\checkmark$&&77.12/58.36&74.96/54.51&74.02/53.68&72.98/52.91&72.01/52.27
\\ %
          &$\checkmark$&77.15/58.70&75.19/54.58&74.31/53.66&73.68/53.07&72.52/52.28
\\ %
$\checkmark$&$\checkmark$&77.83/59.72	&75.74/55.85&74.99/54.82&74.10/54.07&72.89/53.21\\
\midrule
%
%
%
%

    \multirow{2}{*}{FL}&\multirow{2}{*}{OL}& \multicolumn{5}{c}{Latency of Vehicle}  \\
    \cline{3-7}
    &&0ms&100ms&200ms&300ms&400ms\\
    \hline  
        &&76.43/57.74&75.09/52.81&72.84/46.54&69.14/39.12&61.53/29.15
\\
      $\checkmark$&&77.12/58.36&75.69/53.23&72.76/46.85&69.25/39.78&62.13/30.41\\ %
      &$\checkmark$&77.15/58.70&75.80/53.90&73.34/48.49&70.12/41.80&63.21/32.02
\\ %
      $\checkmark$&$\checkmark$&77.83/59.72&76.41/54.88&74.24/49.46&70.75/42.62&63.59/32.31 \\
\bottomrule
\end{tabular}
\begin{tablenotes}
\item FL: \textit{Field loss}. OL: \textit{Offset loss}.
\end{tablenotes}
\end{threeparttable}

\label{tab:ablation_component_loss}
\end{table}

\begin{figure}[t]
    \centering
    \subfloat[LiDAR points.]{\includegraphics[height=2.3in]{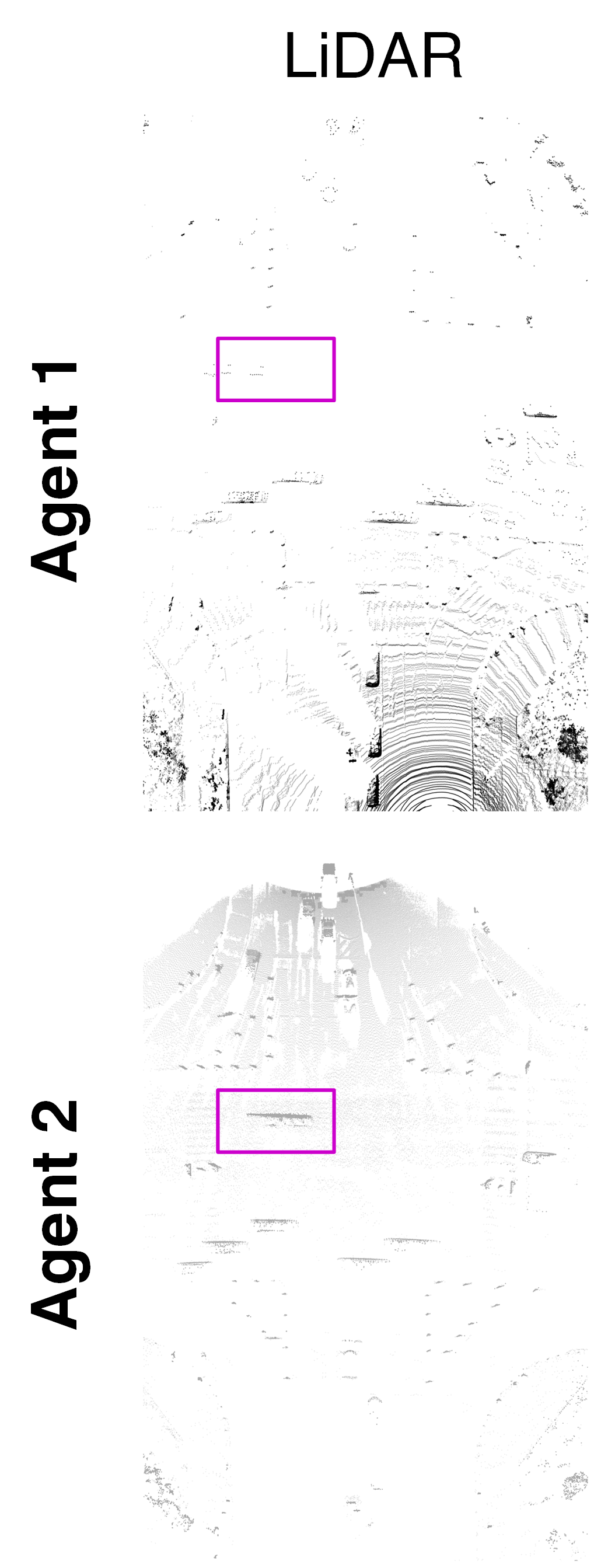}}
    \hspace{1mm}
    \subfloat[Trajectory field.]{\includegraphics[height=2.3in]{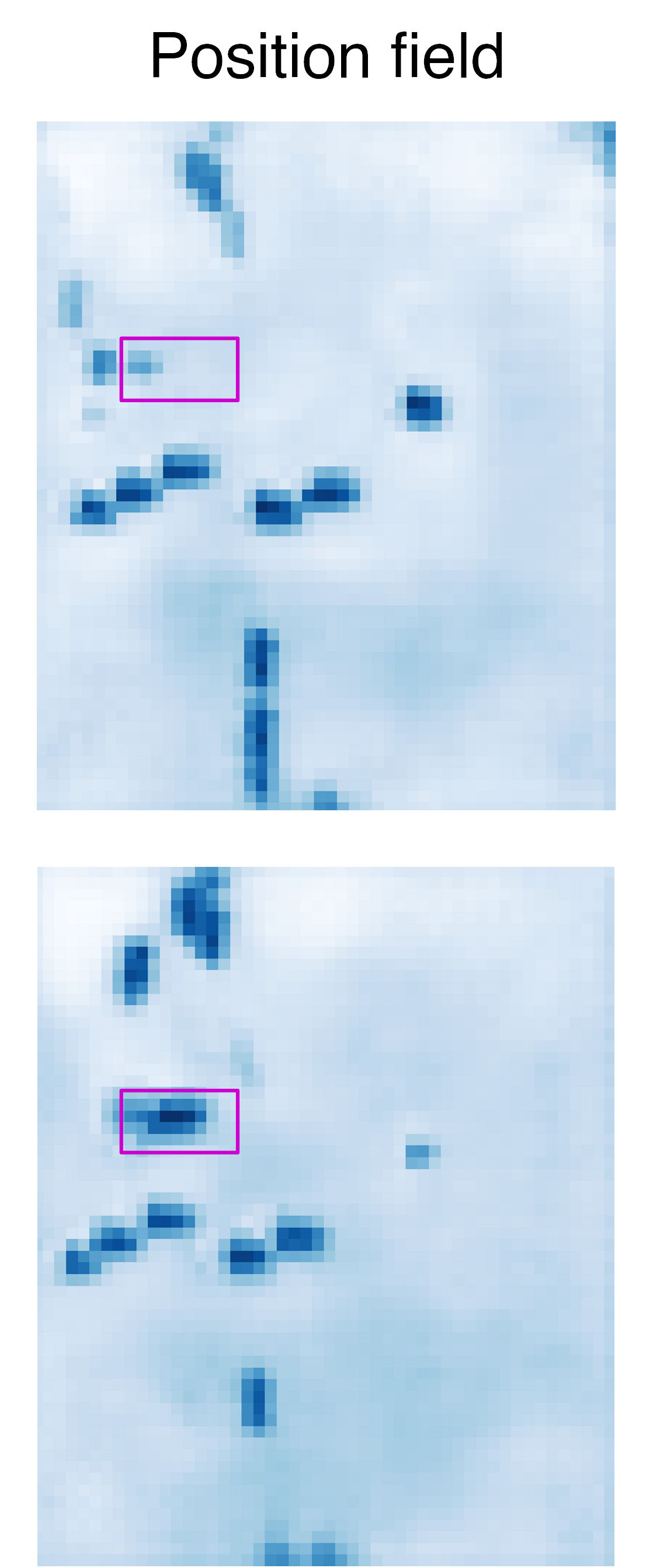}\label{fig:visual_traj_field}}
    \hspace{1mm}
    \subfloat[Deep features.]{\includegraphics[height=2.3in]{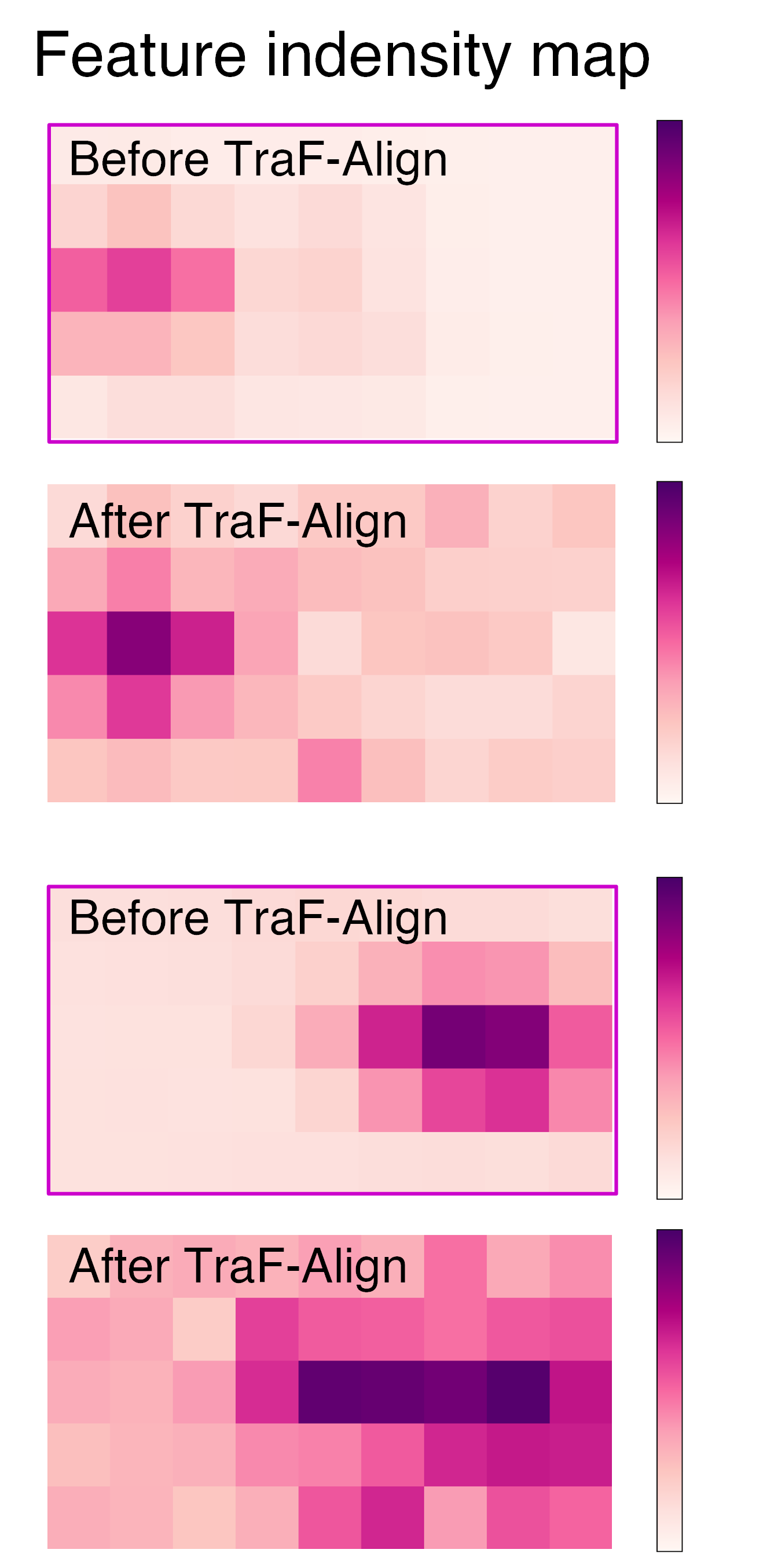}
    \label{fig:visual_deep_defu}}
    \caption{Visualization of \mymethod \ in real-world scenarios from the DAIR-V2X-Seq dataset with latency of $400$ ms.}
    \label{fig:whole_visual}
\end{figure}

\begin{table}[t]
\footnotesize
\centering
\caption{Results of the ablation study of the number of LiDAR frames on the V2V4Real dataset.}
\footnotesize
\definecolor{color}{RGB}{255, 0, 0}   
\renewcommand\arraystretch{1.1} 
\setlength{\tabcolsep}{1pt} 
\renewcommand\arraystretch{1.1} 
\begin{threeparttable}
\begin{tabular}{cc|c|c|c|c|c}
    \toprule
    \multirow{2}{*}{Ego}&\multirow{2}{*}{CV}& \multicolumn{5}{c}{Latency of Cooperative Vehicle (CV)}  \\
    \cline{3-7}
    &&0ms&100ms&200ms&300ms&400ms\\
    \hline  
1&1&73.35/40.54&69.76/34.03&66.51/32.54&65.17/32.13&64.28/32.28\\
4&1&72.85/44.81&69.45/38.32&66.73/37.16&66.00/37.28&65.67/37.32\\
4&2&73.02/\textbf{45.00}&71.74/39.88&69.71/38.25&67.88/37.49&66.98/37.39\\   
1&4&71.28/42.16&70.28/39.18&69.31/38.15&67.68/37.52&66.47/36.90\\
2&4&74.28/44.14&73.04/\textbf{39.99}&71.51/\textbf{39.54}&70.07/\textbf{38.99}&\textbf{69.41}/\textbf{38.49}\\ 
4&4&\textbf{74.76}/44.03&\textbf{73.28}/39.58&\textbf{71.64}/38.58&\textbf{70.10}/37.70&69.08/37.08
\\

\bottomrule
\end{tabular}
\end{threeparttable}


\label{tab:ablation_his_frame}
\end{table}

\begin{figure*}[t]
    \centering
    \subfloat[V2VNet]{
    \label{fig:visual_a}
    \includegraphics[height=1.9in]{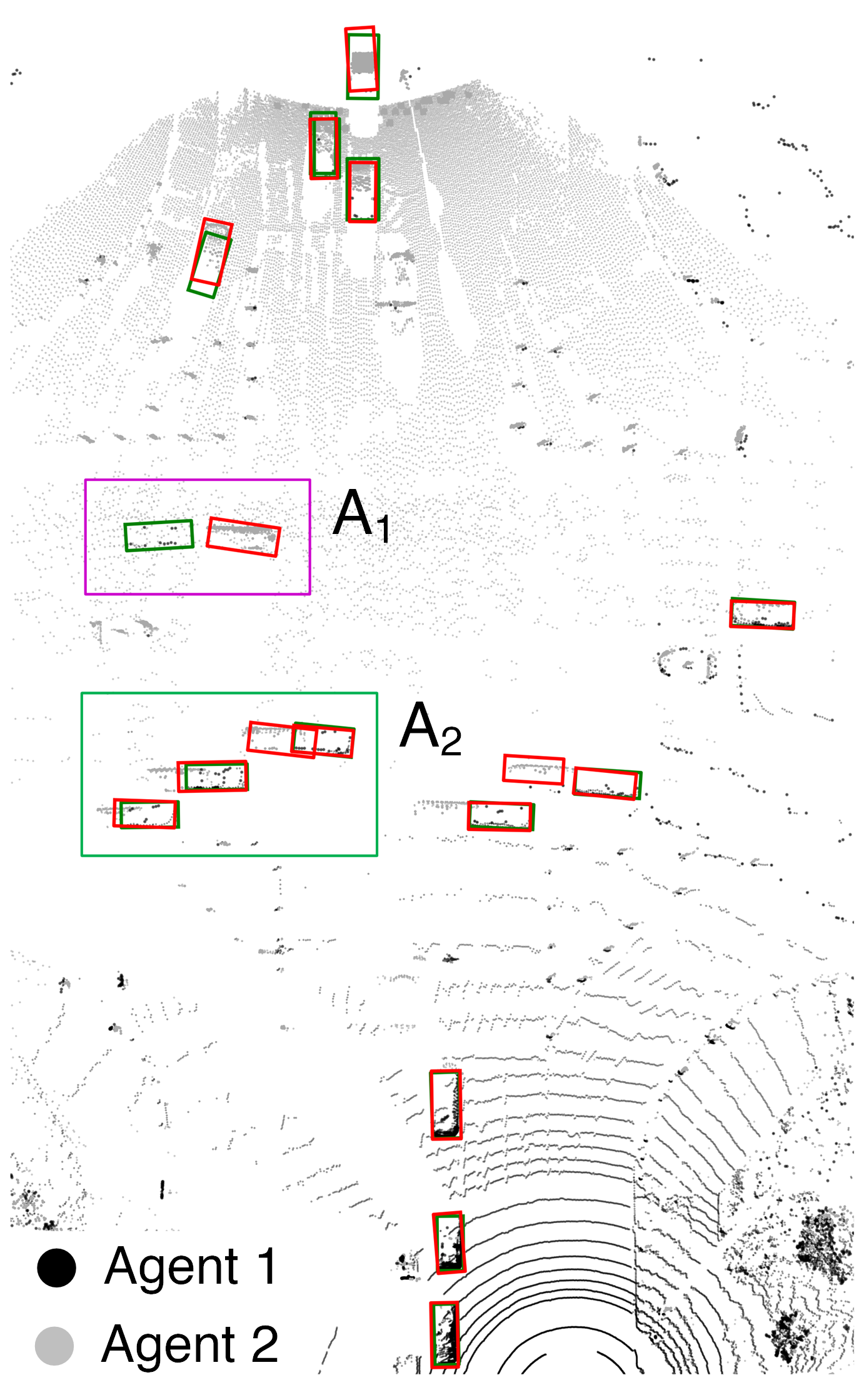}}
    \hspace{-2mm}
    \subfloat[COBEVT]{
    \includegraphics[height=1.9in]{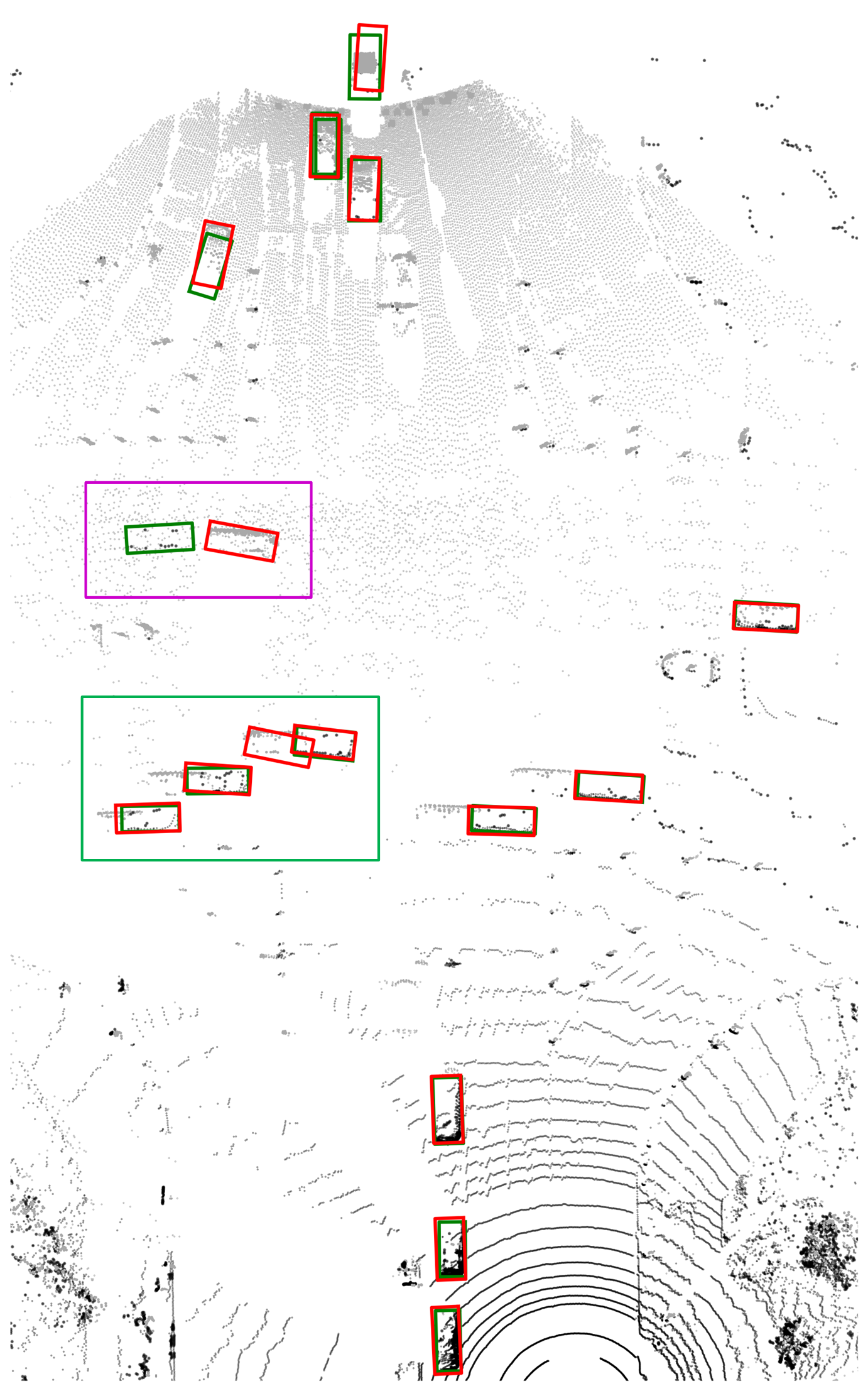}}
    \hspace{-4mm}
    \subfloat[V2X-ViT]{
    \includegraphics[height=1.9in]{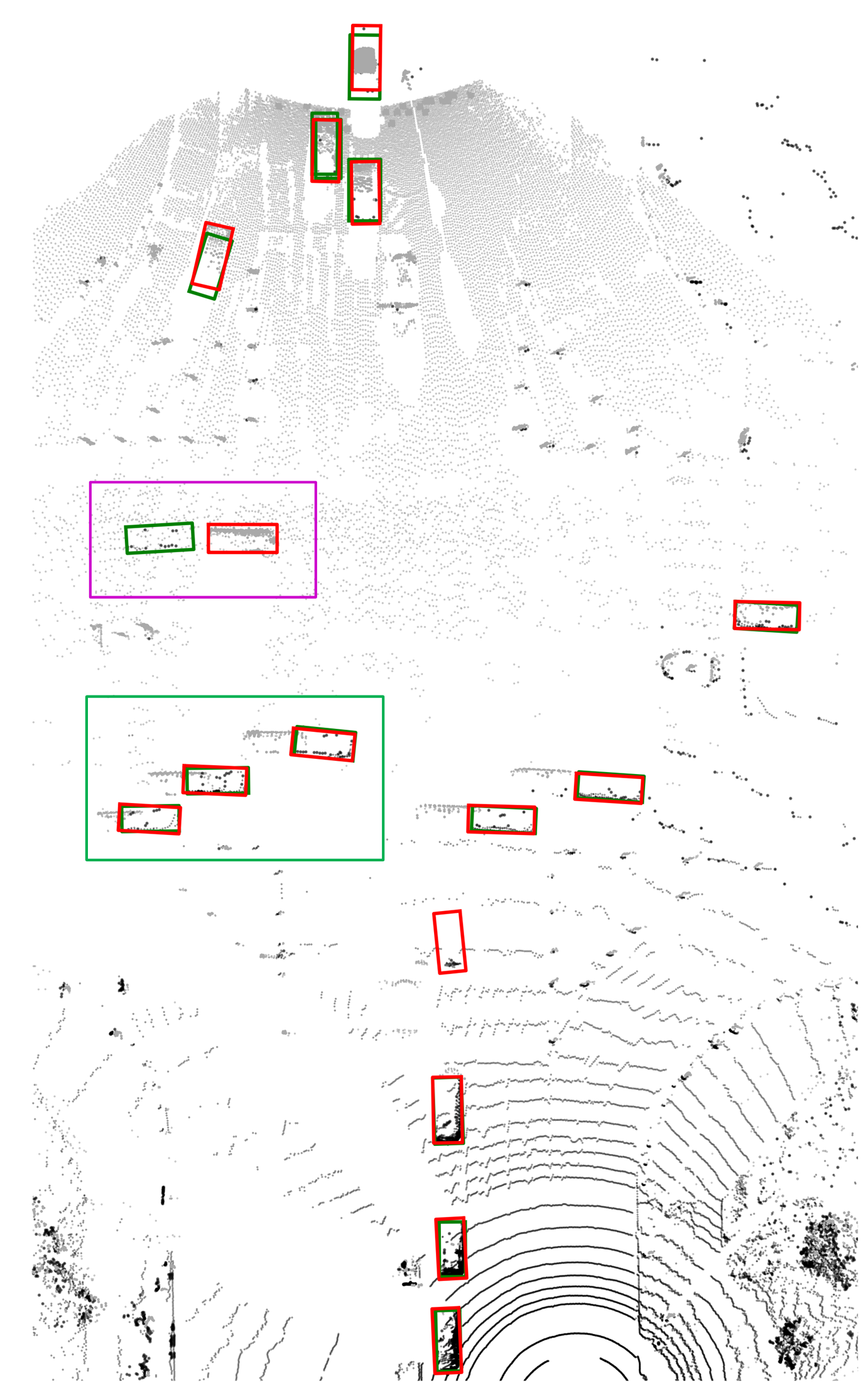}}
    \hspace{-1mm}
    \subfloat[\textbf{\mymethod} (Ours)]{\includegraphics[height=1.9in]{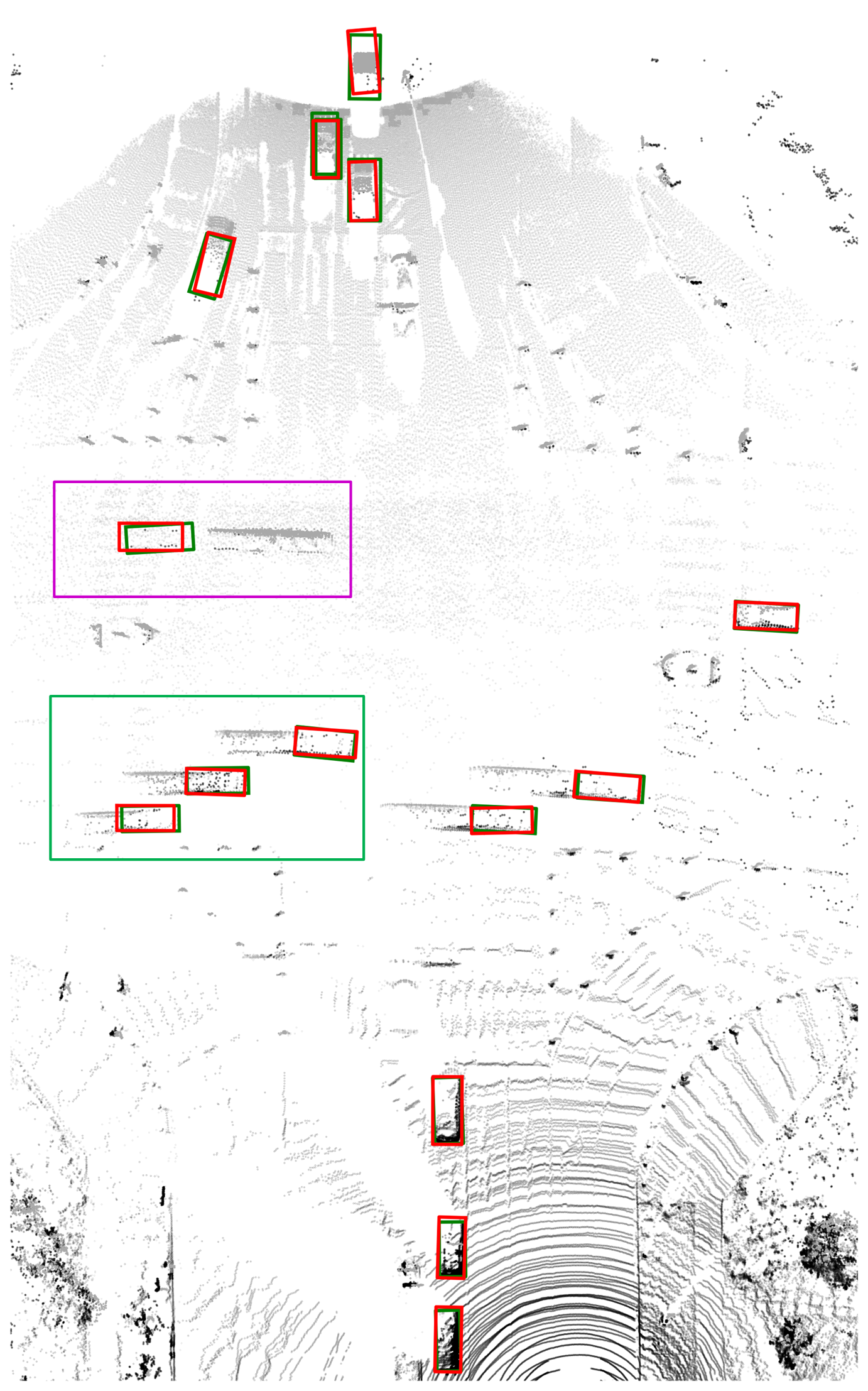}
    \label{fig:visual_defu}}
    \hspace{-0.5mm}
    \subfloat[Learned attention offsets at scale 1/4]{\includegraphics[height=1.9in]{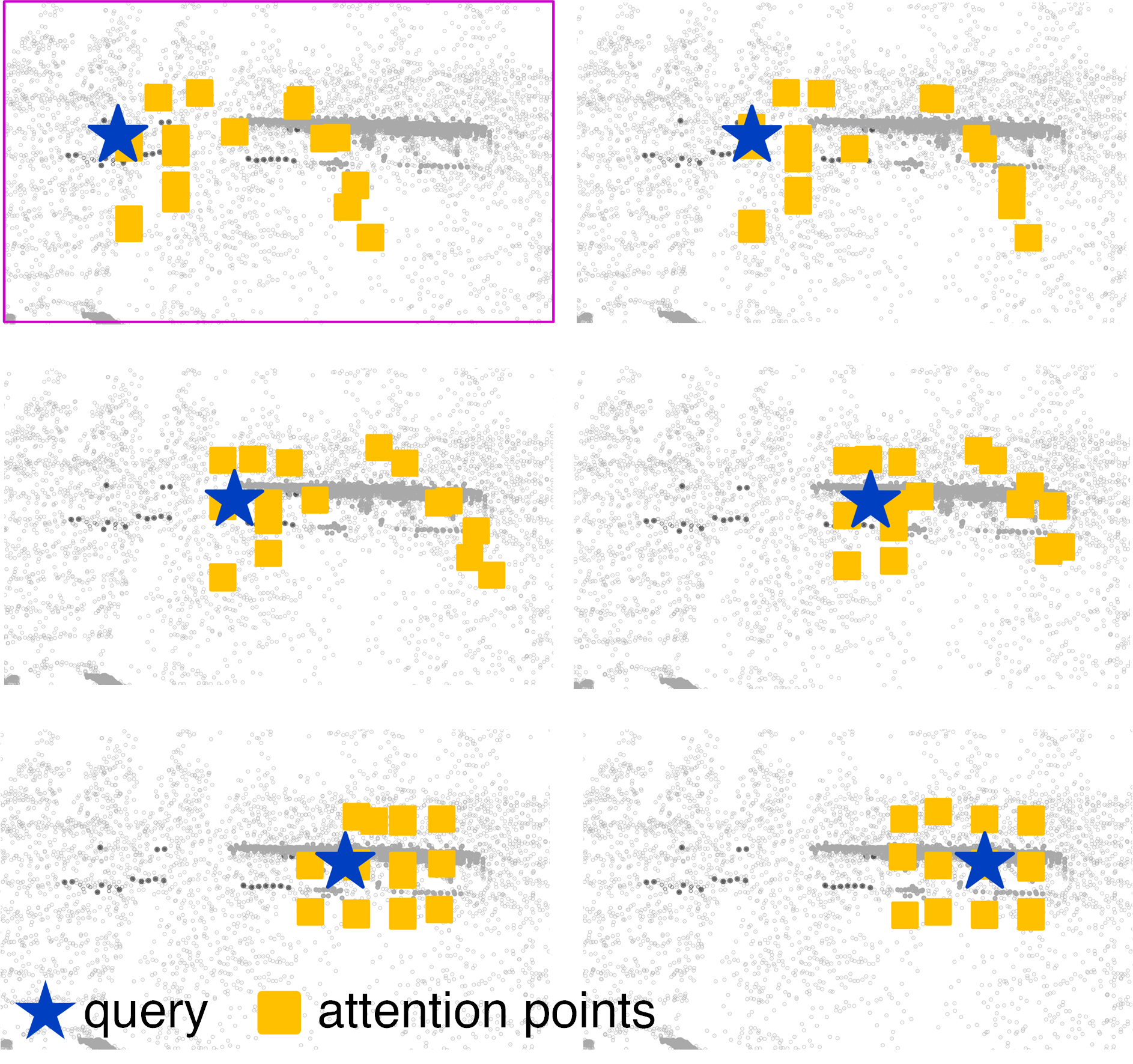}
    \label{fig:visual_defu_att}}
    \caption{Qualitative results in real-world scenarios from the DAIR-V2X-Seq dataset with latency of $400$ ms. Green and red boxes denote the ground truths and detection results, respectively.}
    \label{fig:visualization_sota}
\end{figure*}

\noindent{\textbf{Impact of LiDAR frames.}}  Table \ref{tab:ablation_his_frame} presents the ablation study conducted on V2V4Real, investigating the impact of different historical frame numbers for both the ego and cooperative vehicle on detection performance and robustness to time delays. It's observed that increasing the ego frames primarily enhances AP70, indicating improved accuracy in object localization, while AP50 shows no significant increase.  Conversely, increasing the CV frames enhances both AP50 and the model's robustness to time delays, as most latency originates from the cooperative vehicle's LiDAR. Furthermore, having too many point cloud frames for the ego vehicle may not yield optimal results, as it could introduce additional motion blur issues, which have been proved in the context of single-vehicle perception \cite{rong2023dynstatf}.

\subsection{Qualitative results}
\noindent \textbf{Visualization of detection results.} Figure \ref{fig:visual_a}-\ref{fig:visual_defu} compare \mymethod \ with three previous methods on a real-world scenario from DAIR-V2X-Seq dataset under a latency of $400$ ms.
It is observed that in region $A_1$, due to the latency in roadside (Agent 2), V2VNet, COBEVT, and V2X-ViT all exhibit a false positive detection and fail to capture the correct bounding box; a similar issue occurs in region $A_2$. The reasons for this phenomenon have been discussed in section \ref{sec:intro}. In contrast, \mymethod \ not only identifies the object but also avoids false positives. This improvement is due to our trajectory-guided attention, which enables both agents to achieve a consensus on the same object.

\noindent \textbf{Visualization of trajectory field.} Figure \ref{fig:visual_traj_field} illustrates the predicted trajectory fields (specifically, the position field) from both the vehicle (Agent 1) and roadside (Agent 2) perspectives. Despite the delays in the roadside point clouds, the predicted trajectory field is aligned to the latest frame.

\noindent \textbf{Visualization of attention.} 
Figure \ref{fig:visual_defu_att} shows the response within the $A_1$ box of Figure \ref{fig:visual_defu} from the roadside perspective, with six different query positions. 
Since the ego data is undelayed, the offset distribution centers around the query points, resembling a convolution kernel. In contrast, the roadside agent extends attention along the trajectory.
After fusion by \mymethod, as shown in Figure \ref{fig:visual_deep_defu}, the roadside features align toward the vehicles, demonstrating spatial alignment. The feature intensity near the trajectory is uniform, indicating semantic alignment. More results are available in the supplementary material.

\section{Conclusions}
We propose \mymethod, a novel cooperative perception framework robust to inter-agent latencies. 
Our method handles delays by predicting the flow of object features and aligning them across agents through attention within the expected trajectory paths. 
Extensive experiments on real-world datasets demonstrate its effectiveness.

\noindent \textbf{Limitations and future work.} \mymethod \ is primarily trained on data with regular latencies (multiples of 100 ms). In practical applications, more realistic latency settings should be considered. Additionally, while \mymethod \ is robust to latency, detection precision under delay can still be improved. The main challenges are:  
i) Designing more accurate trajectory field predictors and attention position generators to enhance feature reconstruction capabilities;  
ii) Improving object localization precision. Although multi-frame information improves motion understanding, it also introduces motion blur, making precise object localization challenging and  
iii) Addressing pose estimation errors. Multi-frame processing relies on accurate inter-frame and inter-agent pose alignment, errors in these estimates can degrade performance. Future work will explore methods to reduce the impact of these factors.

\noindent \textbf{Acknowledgments.} This research is supported by the National Key Research and Development Program of China under Grant 2024YFB2505803.

{\small
\bibliographystyle{ieeenat_fullname}
\bibliography{main.bib}
}
\clearpage
\setcounter{section}{0}  
\setcounter{figure}{0}  
\setcounter{table}{0} 
\renewcommand{\thesection}{\arabic{section}} 
\maketitlesupplementary
In this supplementary material, we present additional experimental results, including PR curves from ablation studies, as well as visualizations of the trajectory field and attention positions.

\section{Ablation studies}
\subsection{Major components}
Four combinations of the main components are defined in Table \ref{tab:sup_comp_def}, and their PR-curves are shown in Figure \ref{fig:pr_curve_veh_v2xseq}. Under synchronous conditions, the differences among the ablations are not significant. However, under a $400$ ms delay, \mymethod \ demonstrates a significant advantage across both entire recall ranges. This highlights the positive impact of the modules in improving robustness against delays.

\subsection{Loss}
The PR-curves for different loss combinations are shown in Figure \ref{fig:pr_curve_veh_v2xseq_loss}. As in Figure \ref{fig:pr_curve_veh_v2xseq}, only the curves under synchronous and $400$ ms delay conditions are presented for clarity. The results are consistent with those in Table \ref{tab:ablation_component_loss} of the main text.

\subsection{Historical frames}
Figure \ref{fig:prcurve_frames} presents the Precision-Recall curves for \mymethod \ evaluated on the V2V4Real validation dataset under varying frame settings. For a detailed analysis, please refer to Section \ref{sec:ablation} in the main text.

\begin{table}[b]
\centering
\footnotesize
\caption{Definition of ablations corresponding to Figure \ref{fig:pr_curve_veh_v2xseq}.}
\renewcommand\arraystretch{1.1}
\setlength{\tabcolsep}{4.5pt} 
\renewcommand\arraystretch{1.1}
\begin{tabular}{c|c|c|c}
    \toprule
    Item &Field predictor&Offset generator&Attention layers\\
    \hline  
    Ablation 1 &&&\\%
    Ablation 2&&&$\checkmark$\\
     Ablation 3&&$\checkmark$&$\checkmark$ \\    
 \mymethod & $\checkmark$&$\checkmark$&$\checkmark$ \\
\bottomrule
\end{tabular}
\label{tab:sup_comp_def}
\end{table}

\begin{figure}[t]
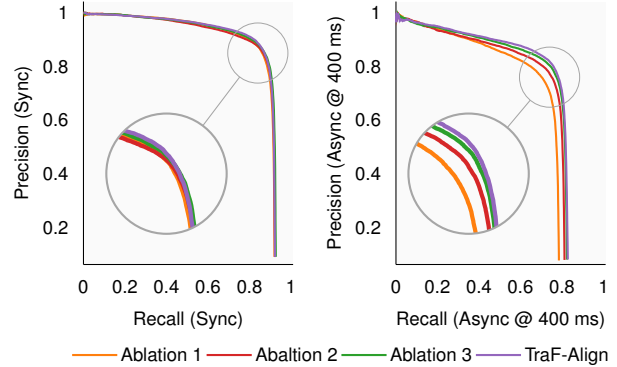

\centering
\subfloat{
\input{table/ablation/pr_curve/component_inf_0ms}}
\subfloat{
\input{table/ablation/pr_curve/component_inf_400ms}}
\vfill
\subfloat{
\definecolor{crimson2143940}{RGB}{214,39,40}
\definecolor{darkgray176}{RGB}{176,176,176}
\definecolor{darkorange25512714}{RGB}{255,127,14}
\definecolor{forestgreen4416044}{RGB}{44,160,44}
\definecolor{lightgray204}{RGB}{204,204,204}
\definecolor{mediumpurple148103189}{RGB}{148,103,189}
\definecolor{steelblue31119180}{RGB}{31,119,180}

\begin{tikzpicture}[font=\sffamily\scriptsize]
\node[draw=white,outer sep=0.5pt,inner sep=0pt] at (0,0) 
{
    \renewcommand{\arraystretch}{1.25} 
    \setlength{\tabcolsep}{0.5pt} 
     \begin{tabular}{cccccccc}
     \ref{pr_curve_com_Ablation 1}&Ablation 1 &
     \ref{pr_curve_com_Ablation 2} & Abaltion 2&\ref{pr_curve_com_Ablation 3} & Ablation 3&
      \ref{pr_curve_com_mymethod} &\mymethod \\
    \end{tabular}
    };
\end{tikzpicture}}
\caption{Precision-Recall curve showing the ablation results of major components when the Ego vehicle experiences $0$ ms and $400$ ms delays on the DAIR-V2X-Seq dataset.}
\label{fig:pr_curve_veh_v2xseq}
\end{figure}

\begin{figure}[t]
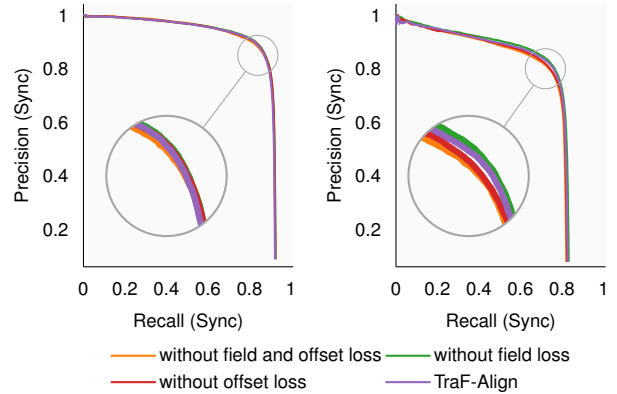

\centering
\subfloat{
\input{table/ablation/pr_curve/loss_inf_0ms}}
\subfloat{
\input{table/ablation/pr_curve/loss_inf_400ms}}
\vfill
\subfloat{
\definecolor{crimson2143940}{RGB}{214,39,40}
\definecolor{darkgray176}{RGB}{176,176,176}
\definecolor{darkorange25512714}{RGB}{255,127,14}
\definecolor{forestgreen4416044}{RGB}{44,160,44}
\definecolor{lightgray204}{RGB}{204,204,204}
\definecolor{mediumpurple148103189}{RGB}{148,103,189}
\definecolor{steelblue31119180}{RGB}{31,119,180}

\begin{tikzpicture}[font=\sffamily\scriptsize]
\node[draw=white,outer sep=0.5pt,inner sep=0pt] at (0,0) 
{
    \renewcommand{\arraystretch}{1.25} 
    \setlength{\tabcolsep}{0.5pt} 
     \begin{tabular}{llll}
     \ref{prcurve_loss_wofsos} &without field and offset loss&
     \ref{prcurve_loss_wofs}&without field loss \\
     \ref{prcurve_loss_woos} & without offset loss& 
      \ref{prcurve_loss_full} & \mymethod \\
    \end{tabular}
    };
\end{tikzpicture}}
\caption{Precision-Recall curve showing the ablation results of loss when the Ego vehicle experiences $0$ ms and $400$ ms delays on the DAIR-V2X-Seq dataset.}
\label{fig:pr_curve_veh_v2xseq_loss}
\end{figure}

\begin{figure*}[t]
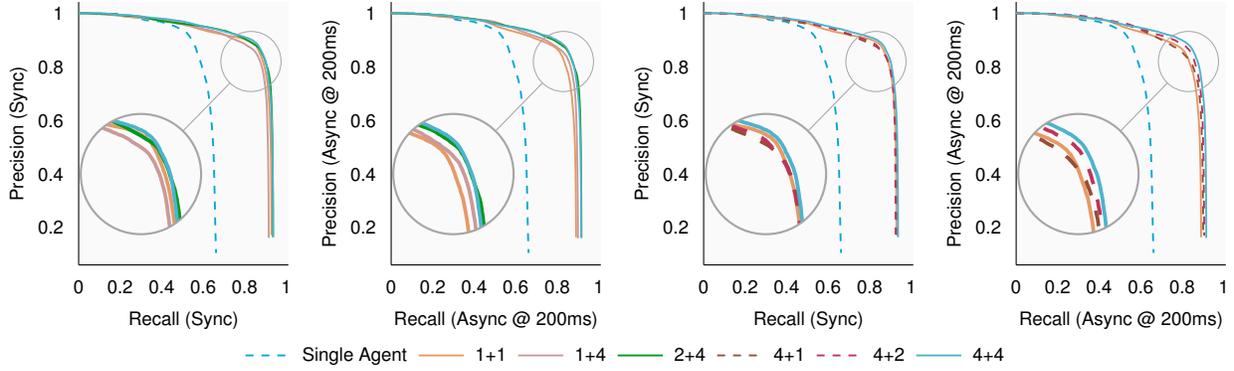

\centering
\subfloat{
\input{table/ablation/pr_curve/frame_0ms_iou0.5_cavframe4}}
\subfloat{
\input{table/ablation/pr_curve/frame_200ms_iou0.5_cavframe4}}
\subfloat{
\input{table/ablation/pr_curve/frame_0ms_iou0.5_egoframe4}}
\subfloat{
\input{table/ablation/pr_curve/frame_200ms_iou0.5_egoframe4}}
\vfill
\subfloat{
\definecolor{mycolor1}{rgb}{0.00000,0.65000,0.85000}
\definecolor{mycolor2}{rgb}{0.95000,0.60000,0.25000}
\definecolor{mycolor3}{rgb}{0.97000,0.82000,0.35000}
\definecolor{mycolor4}{rgb}{0.60000,0.30000,0.75000}
\definecolor{mycolor5}{rgb}{0,0.6,0.1}
\definecolor{mycolor6}{rgb}{0.45000,0.85000,0.95000}
\definecolor{mycolor7}{rgb}{0.75000,0.20000,0.35000}
\definecolor{mycolor8}{rgb}{0.8,0.6,0.6}
\definecolor{mycolor9}{rgb}{0.4,0.7,0.9}
\definecolor{mycolor10}{rgb}{0.7,0.6,0.4}
\definecolor{mycolor11}{rgb}{0.9,0.8,0.3}
\definecolor{mycolor12}{rgb}{0.5,0.4,0.7}
\definecolor{mycolor13}{rgb}{0.6,0.3,0.2}
\definecolor{mycolor14}{rgb}{0.9,0.6,0.4}
\definecolor{mycolor15}{rgb}{0.3,0.7,0.8}

\begin{tikzpicture}[font=\sffamily\scriptsize]
\node[draw=white,outer sep=0.5pt,inner sep=0pt] at (0,0) 
{
    \renewcommand{\arraystretch}{1.25} 
    \setlength{\tabcolsep}{2pt} 
     \begin{tabular}{cccccccccccccc}
     \ref{prcurve_nofusion}&Single Agent &
     \ref{frame_1_1} & 1+1&\ref{frame_1_4} & 1+4&
      \ref{frame_2_4} & 2+4 &
      \ref{frame_4_1} & 4+1&
      \ref{frame_4_2} & 4+2 &  \ref{frame_4_4} & 4+4 \\
    \end{tabular}
    };
\end{tikzpicture}}
\caption{Precision-Recall curves for \mymethod \ on the V2V4Real validation dataset under different frame settings. The legend notation $i+j$ represents the use of $i$ frame(s) for the ego vehicle and $j$ frame(s) for the cooperative vehicle.}
\label{fig:prcurve_frames}
\end{figure*}

\begin{figure*}[t]
    \centering
    \includegraphics[width=0.92\linewidth]{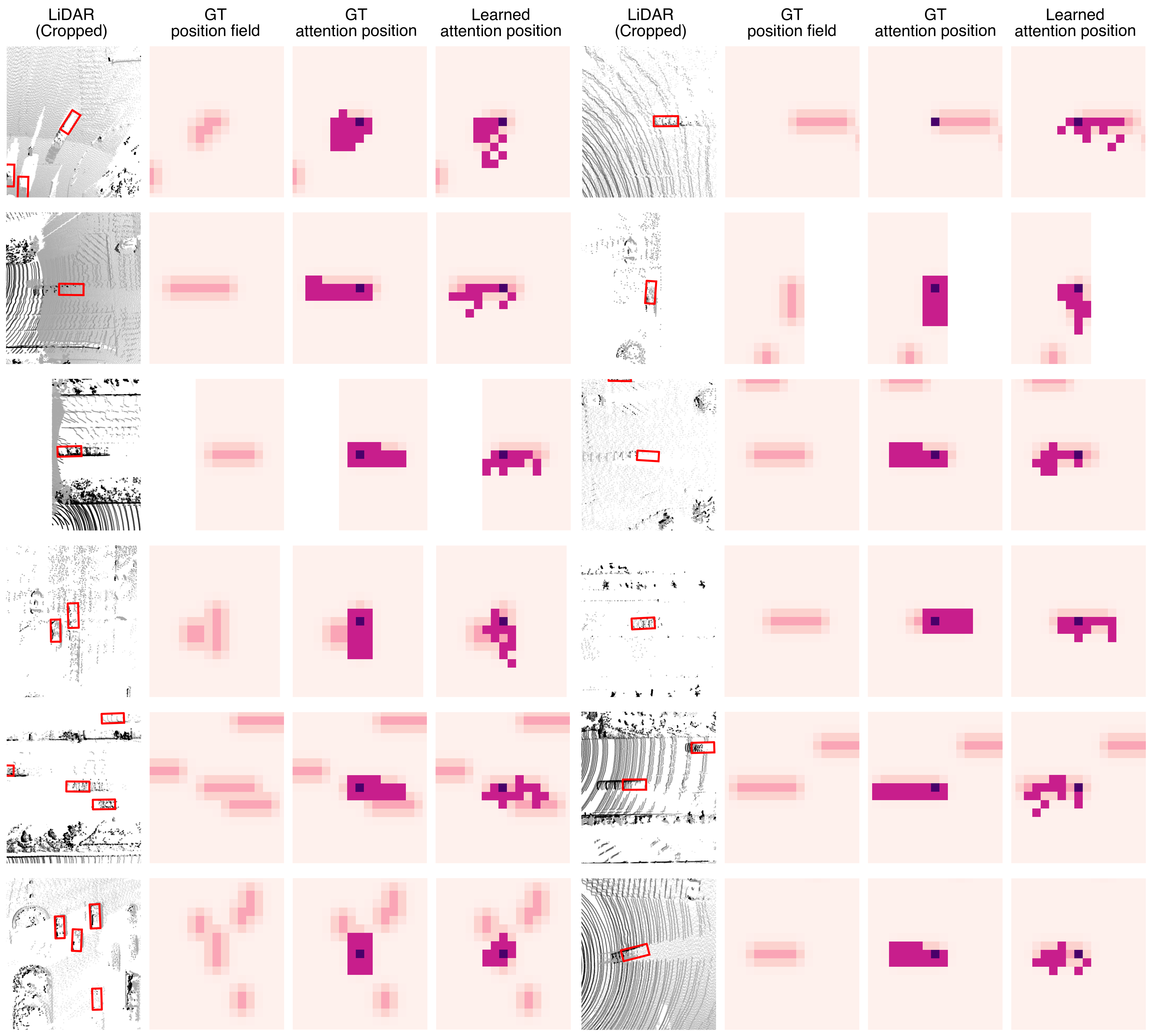}
    \caption{Examples of learned attention positions.}
    \label{fig:supp_offset_visual}
\end{figure*}

\section{Trajectory field} 
\begin{figure*}[t]
    \centering
    \includegraphics[width=0.87\linewidth]{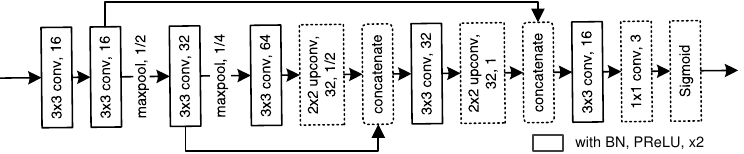}
    \caption{Field predictor.}
    \label{fig:network_fp}
\end{figure*}

\noindent \textbf{Network.} The network for the field predictor (Figure \ref{fig:network_fp}) is adapted from UNet. It first reduces the dimensionality of the input feature map, then progressively upsamples it while concatenating the input feature map at each stage. During this process, as the feature map is downsampled, deeper semantic information is progressively extracted. In the upsampling phase, these features are decoded, allowing for a more comprehensive semantic understanding of the feature map.
In this paper, the field predictor is used to interpret the feature map output from the backbone and generate the trajectory field, which aligns perfectly with this process.

\begin{figure*}
    \centering
    \includegraphics[width=0.95\linewidth]{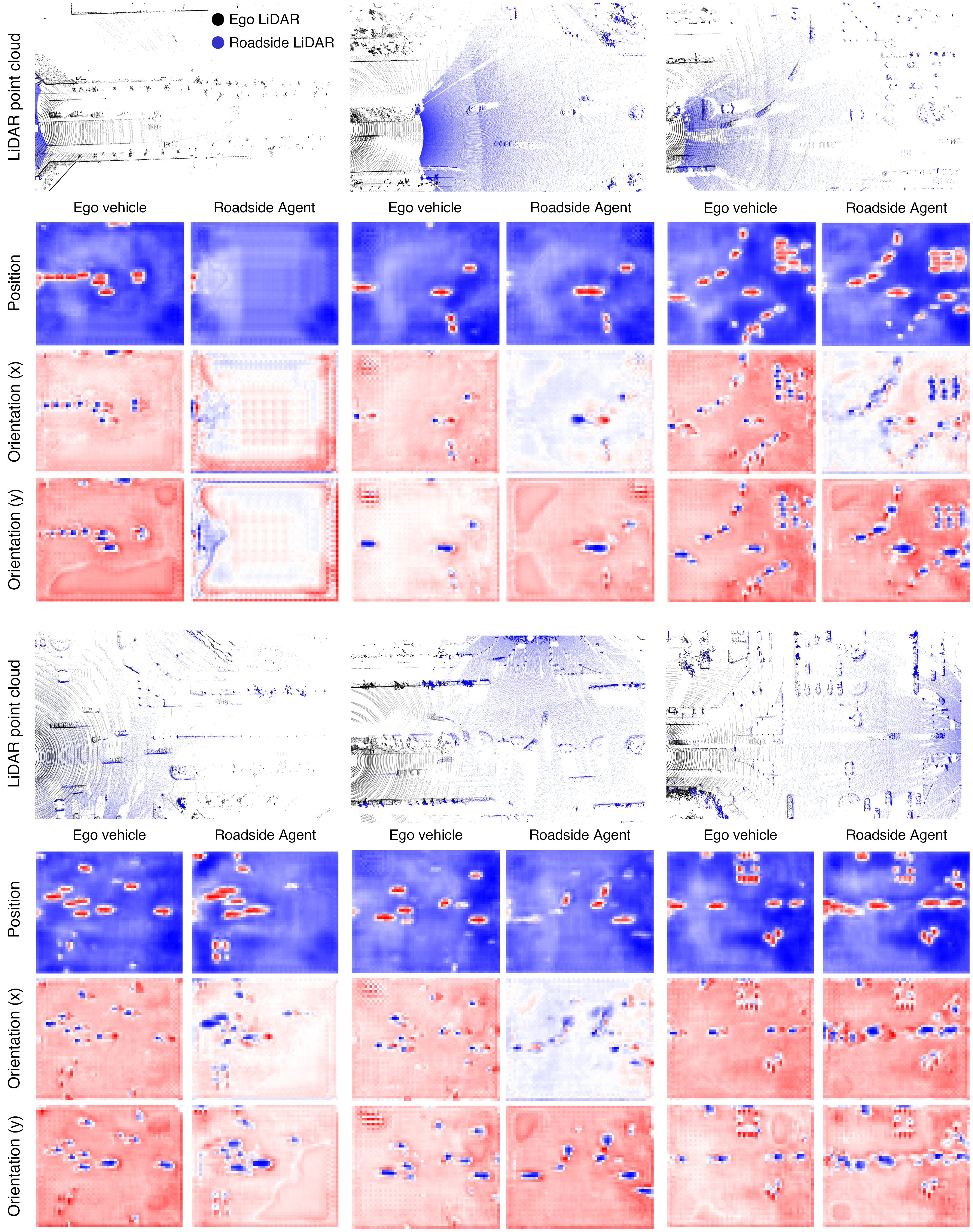}
    \caption{Examples of learned trajectory field by \mymethod \ on DAIR-V2X-Seq dataset \cite{yu2023v2x}. The position field creates heatmap peaks along the object’s trajectory, while the orientation field captures the inverse tangent direction of the trajectory.}
    \label{fig:supp_field}
\end{figure*}

\noindent \textbf{Field visualization.}
Trajectory field ($\mathbb{R}^{3\times H \times W}$) consists of a position field ($\mathbb{R}^{1\times H \times W}$) and an orientation field ($\mathbb{R}^{2\times H \times W}$). The position field creates heatmap peaks along the object’s trajectory, while the orientation field captures the inverse tangent direction of the trajectory. Figure \ref{fig:supp_field} shows several examples of the learned trajectory field.

\begin{figure*}[t]
    \centering
    \includegraphics[width=0.85\linewidth]{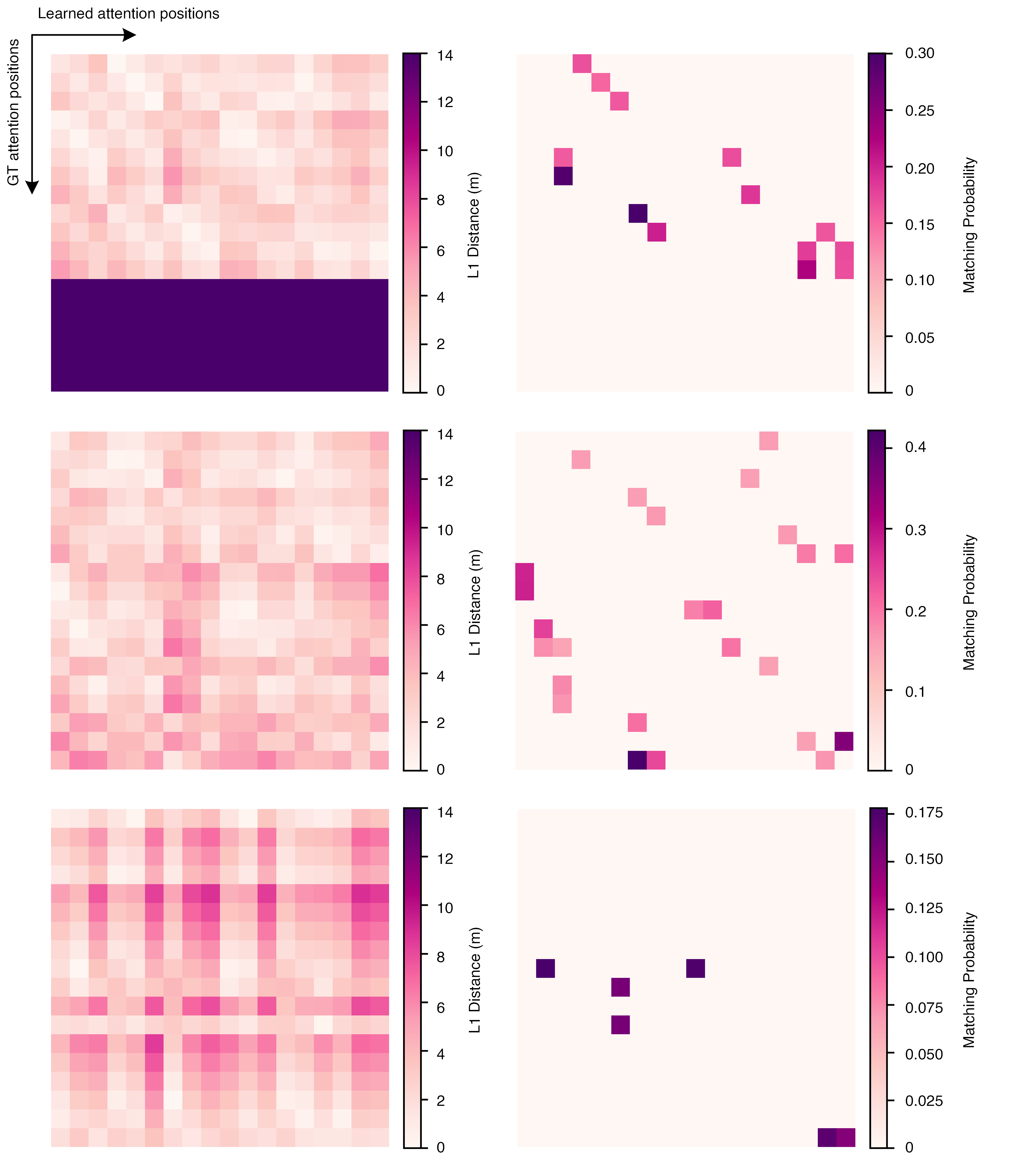}
    \caption{Examples of the cost and assignment probability matrices.}
    \label{fig:assign_matrix}
\end{figure*}

\section{Attention position}
Figure \ref{fig:supp_offset_visual} visualizes attention positions learned by \mymethod, and Figure \ref{fig:assign_matrix} shows Sinkhorn's cost and matching probability matrices.

\end{document}